\documentclass[letterpaper]{article} 
\usepackage{aaai18}  
\usepackage{times}  
\usepackage{helvet} 
\usepackage{courier}
\usepackage[draft]{minted}  
\makeatletter
\def\PYGdefault@reset{\let\PYGdefault@it=\relax \let\PYGdefault@bf=\relax%
    \let\PYGdefault@ul=\relax \let\PYGdefault@tc=\relax%
    \let\PYGdefault@bc=\relax \let\PYGdefault@ff=\relax}
\def\PYGdefault@tok#1{\csname PYGdefault@tok@#1\endcsname}
\def\PYGdefault@toks#1+{\ifx\relax#1\empty\else%
    \PYGdefault@tok{#1}\expandafter\PYGdefault@toks\fi}
\def\PYGdefault@do#1{\PYGdefault@bc{\PYGdefault@tc{\PYGdefault@ul{%
    \PYGdefault@it{\PYGdefault@bf{\PYGdefault@ff{#1}}}}}}}
\def\PYGdefault#1#2{\PYGdefault@reset\PYGdefault@toks#1+\relax+\PYGdefault@do{#2}}

\expandafter\def\csname PYGdefault@tok@gd\endcsname{\def\PYGdefault@tc##1{\textcolor[rgb]{0.63,0.00,0.00}{##1}}}
\expandafter\def\csname PYGdefault@tok@gu\endcsname{\let\PYGdefault@bf=\textbf\def\PYGdefault@tc##1{\textcolor[rgb]{0.50,0.00,0.50}{##1}}}
\expandafter\def\csname PYGdefault@tok@gt\endcsname{\def\PYGdefault@tc##1{\textcolor[rgb]{0.00,0.27,0.87}{##1}}}
\expandafter\def\csname PYGdefault@tok@gs\endcsname{\let\PYGdefault@bf=\textbf}
\expandafter\def\csname PYGdefault@tok@gr\endcsname{\def\PYGdefault@tc##1{\textcolor[rgb]{1.00,0.00,0.00}{##1}}}
\expandafter\def\csname PYGdefault@tok@cm\endcsname{\let\PYGdefault@it=\textit\def\PYGdefault@tc##1{\textcolor[rgb]{0.25,0.50,0.50}{##1}}}
\expandafter\def\csname PYGdefault@tok@vg\endcsname{\def\PYGdefault@tc##1{\textcolor[rgb]{0.10,0.09,0.49}{##1}}}
\expandafter\def\csname PYGdefault@tok@vi\endcsname{\def\PYGdefault@tc##1{\textcolor[rgb]{0.10,0.09,0.49}{##1}}}
\expandafter\def\csname PYGdefault@tok@mh\endcsname{\def\PYGdefault@tc##1{\textcolor[rgb]{0.40,0.40,0.40}{##1}}}
\expandafter\def\csname PYGdefault@tok@cs\endcsname{\let\PYGdefault@it=\textit\def\PYGdefault@tc##1{\textcolor[rgb]{0.25,0.50,0.50}{##1}}}
\expandafter\def\csname PYGdefault@tok@ge\endcsname{\let\PYGdefault@it=\textit}
\expandafter\def\csname PYGdefault@tok@vc\endcsname{\def\PYGdefault@tc##1{\textcolor[rgb]{0.10,0.09,0.49}{##1}}}
\expandafter\def\csname PYGdefault@tok@il\endcsname{\def\PYGdefault@tc##1{\textcolor[rgb]{0.40,0.40,0.40}{##1}}}
\expandafter\def\csname PYGdefault@tok@go\endcsname{\def\PYGdefault@tc##1{\textcolor[rgb]{0.53,0.53,0.53}{##1}}}
\expandafter\def\csname PYGdefault@tok@cp\endcsname{\def\PYGdefault@tc##1{\textcolor[rgb]{0.74,0.48,0.00}{##1}}}
\expandafter\def\csname PYGdefault@tok@gi\endcsname{\def\PYGdefault@tc##1{\textcolor[rgb]{0.00,0.63,0.00}{##1}}}
\expandafter\def\csname PYGdefault@tok@gh\endcsname{\let\PYGdefault@bf=\textbf\def\PYGdefault@tc##1{\textcolor[rgb]{0.00,0.00,0.50}{##1}}}
\expandafter\def\csname PYGdefault@tok@ni\endcsname{\let\PYGdefault@bf=\textbf\def\PYGdefault@tc##1{\textcolor[rgb]{0.60,0.60,0.60}{##1}}}
\expandafter\def\csname PYGdefault@tok@nl\endcsname{\def\PYGdefault@tc##1{\textcolor[rgb]{0.63,0.63,0.00}{##1}}}
\expandafter\def\csname PYGdefault@tok@nn\endcsname{\let\PYGdefault@bf=\textbf\def\PYGdefault@tc##1{\textcolor[rgb]{0.00,0.00,1.00}{##1}}}
\expandafter\def\csname PYGdefault@tok@no\endcsname{\def\PYGdefault@tc##1{\textcolor[rgb]{0.53,0.00,0.00}{##1}}}
\expandafter\def\csname PYGdefault@tok@na\endcsname{\def\PYGdefault@tc##1{\textcolor[rgb]{0.49,0.56,0.16}{##1}}}
\expandafter\def\csname PYGdefault@tok@nb\endcsname{\def\PYGdefault@tc##1{\textcolor[rgb]{0.00,0.50,0.00}{##1}}}
\expandafter\def\csname PYGdefault@tok@nc\endcsname{\let\PYGdefault@bf=\textbf\def\PYGdefault@tc##1{\textcolor[rgb]{0.00,0.00,1.00}{##1}}}
\expandafter\def\csname PYGdefault@tok@nd\endcsname{\def\PYGdefault@tc##1{\textcolor[rgb]{0.67,0.13,1.00}{##1}}}
\expandafter\def\csname PYGdefault@tok@ne\endcsname{\let\PYGdefault@bf=\textbf\def\PYGdefault@tc##1{\textcolor[rgb]{0.82,0.25,0.23}{##1}}}
\expandafter\def\csname PYGdefault@tok@nf\endcsname{\def\PYGdefault@tc##1{\textcolor[rgb]{0.00,0.00,1.00}{##1}}}
\expandafter\def\csname PYGdefault@tok@si\endcsname{\let\PYGdefault@bf=\textbf\def\PYGdefault@tc##1{\textcolor[rgb]{0.73,0.40,0.53}{##1}}}
\expandafter\def\csname PYGdefault@tok@s2\endcsname{\def\PYGdefault@tc##1{\textcolor[rgb]{0.73,0.13,0.13}{##1}}}
\expandafter\def\csname PYGdefault@tok@nt\endcsname{\let\PYGdefault@bf=\textbf\def\PYGdefault@tc##1{\textcolor[rgb]{0.00,0.50,0.00}{##1}}}
\expandafter\def\csname PYGdefault@tok@nv\endcsname{\def\PYGdefault@tc##1{\textcolor[rgb]{0.10,0.09,0.49}{##1}}}
\expandafter\def\csname PYGdefault@tok@s1\endcsname{\def\PYGdefault@tc##1{\textcolor[rgb]{0.73,0.13,0.13}{##1}}}
\expandafter\def\csname PYGdefault@tok@ch\endcsname{\let\PYGdefault@it=\textit\def\PYGdefault@tc##1{\textcolor[rgb]{0.25,0.50,0.50}{##1}}}
\expandafter\def\csname PYGdefault@tok@m\endcsname{\def\PYGdefault@tc##1{\textcolor[rgb]{0.40,0.40,0.40}{##1}}}
\expandafter\def\csname PYGdefault@tok@gp\endcsname{\let\PYGdefault@bf=\textbf\def\PYGdefault@tc##1{\textcolor[rgb]{0.00,0.00,0.50}{##1}}}
\expandafter\def\csname PYGdefault@tok@sh\endcsname{\def\PYGdefault@tc##1{\textcolor[rgb]{0.73,0.13,0.13}{##1}}}
\expandafter\def\csname PYGdefault@tok@ow\endcsname{\let\PYGdefault@bf=\textbf\def\PYGdefault@tc##1{\textcolor[rgb]{0.67,0.13,1.00}{##1}}}
\expandafter\def\csname PYGdefault@tok@sx\endcsname{\def\PYGdefault@tc##1{\textcolor[rgb]{0.00,0.50,0.00}{##1}}}
\expandafter\def\csname PYGdefault@tok@bp\endcsname{\def\PYGdefault@tc##1{\textcolor[rgb]{0.00,0.50,0.00}{##1}}}
\expandafter\def\csname PYGdefault@tok@c1\endcsname{\let\PYGdefault@it=\textit\def\PYGdefault@tc##1{\textcolor[rgb]{0.25,0.50,0.50}{##1}}}
\expandafter\def\csname PYGdefault@tok@o\endcsname{\def\PYGdefault@tc##1{\textcolor[rgb]{0.40,0.40,0.40}{##1}}}
\expandafter\def\csname PYGdefault@tok@kc\endcsname{\let\PYGdefault@bf=\textbf\def\PYGdefault@tc##1{\textcolor[rgb]{0.00,0.50,0.00}{##1}}}
\expandafter\def\csname PYGdefault@tok@c\endcsname{\let\PYGdefault@it=\textit\def\PYGdefault@tc##1{\textcolor[rgb]{0.25,0.50,0.50}{##1}}}
\expandafter\def\csname PYGdefault@tok@mf\endcsname{\def\PYGdefault@tc##1{\textcolor[rgb]{0.40,0.40,0.40}{##1}}}
\expandafter\def\csname PYGdefault@tok@err\endcsname{\def\PYGdefault@bc##1{\setlength{\fboxsep}{0pt}\fcolorbox[rgb]{1.00,0.00,0.00}{1,1,1}{\strut ##1}}}
\expandafter\def\csname PYGdefault@tok@mb\endcsname{\def\PYGdefault@tc##1{\textcolor[rgb]{0.40,0.40,0.40}{##1}}}
\expandafter\def\csname PYGdefault@tok@ss\endcsname{\def\PYGdefault@tc##1{\textcolor[rgb]{0.10,0.09,0.49}{##1}}}
\expandafter\def\csname PYGdefault@tok@sr\endcsname{\def\PYGdefault@tc##1{\textcolor[rgb]{0.73,0.40,0.53}{##1}}}
\expandafter\def\csname PYGdefault@tok@mo\endcsname{\def\PYGdefault@tc##1{\textcolor[rgb]{0.40,0.40,0.40}{##1}}}
\expandafter\def\csname PYGdefault@tok@kd\endcsname{\let\PYGdefault@bf=\textbf\def\PYGdefault@tc##1{\textcolor[rgb]{0.00,0.50,0.00}{##1}}}
\expandafter\def\csname PYGdefault@tok@mi\endcsname{\def\PYGdefault@tc##1{\textcolor[rgb]{0.40,0.40,0.40}{##1}}}
\expandafter\def\csname PYGdefault@tok@kn\endcsname{\let\PYGdefault@bf=\textbf\def\PYGdefault@tc##1{\textcolor[rgb]{0.00,0.50,0.00}{##1}}}
\expandafter\def\csname PYGdefault@tok@cpf\endcsname{\let\PYGdefault@it=\textit\def\PYGdefault@tc##1{\textcolor[rgb]{0.25,0.50,0.50}{##1}}}
\expandafter\def\csname PYGdefault@tok@kr\endcsname{\let\PYGdefault@bf=\textbf\def\PYGdefault@tc##1{\textcolor[rgb]{0.00,0.50,0.00}{##1}}}
\expandafter\def\csname PYGdefault@tok@s\endcsname{\def\PYGdefault@tc##1{\textcolor[rgb]{0.73,0.13,0.13}{##1}}}
\expandafter\def\csname PYGdefault@tok@kp\endcsname{\def\PYGdefault@tc##1{\textcolor[rgb]{0.00,0.50,0.00}{##1}}}
\expandafter\def\csname PYGdefault@tok@w\endcsname{\def\PYGdefault@tc##1{\textcolor[rgb]{0.73,0.73,0.73}{##1}}}
\expandafter\def\csname PYGdefault@tok@kt\endcsname{\def\PYGdefault@tc##1{\textcolor[rgb]{0.69,0.00,0.25}{##1}}}
\expandafter\def\csname PYGdefault@tok@sc\endcsname{\def\PYGdefault@tc##1{\textcolor[rgb]{0.73,0.13,0.13}{##1}}}
\expandafter\def\csname PYGdefault@tok@sb\endcsname{\def\PYGdefault@tc##1{\textcolor[rgb]{0.73,0.13,0.13}{##1}}}
\expandafter\def\csname PYGdefault@tok@k\endcsname{\let\PYGdefault@bf=\textbf\def\PYGdefault@tc##1{\textcolor[rgb]{0.00,0.50,0.00}{##1}}}
\expandafter\def\csname PYGdefault@tok@se\endcsname{\let\PYGdefault@bf=\textbf\def\PYGdefault@tc##1{\textcolor[rgb]{0.73,0.40,0.13}{##1}}}
\expandafter\def\csname PYGdefault@tok@sd\endcsname{\let\PYGdefault@it=\textit\def\PYGdefault@tc##1{\textcolor[rgb]{0.73,0.13,0.13}{##1}}}

\makeatother
\makeatletter
\def\PYG@reset{\let\PYG@it=\relax \let\PYG@bf=\relax%
    \let\PYG@ul=\relax \let\PYG@tc=\relax%
    \let\PYG@bc=\relax \let\PYG@ff=\relax}
\def\PYG@tok#1{\csname PYG@tok@#1\endcsname}
\def\PYG@toks#1+{\ifx\relax#1\empty\else%
    \PYG@tok{#1}\expandafter\PYG@toks\fi}
\def\PYG@do#1{\PYG@bc{\PYG@tc{\PYG@ul{%
    \PYG@it{\PYG@bf{\PYG@ff{#1}}}}}}}
\def\PYG#1#2{\PYG@reset\PYG@toks#1+\relax+\PYG@do{#2}}

\expandafter\def\csname PYG@tok@gd\endcsname{\def\PYG@tc##1{\textcolor[rgb]{0.63,0.00,0.00}{##1}}}
\expandafter\def\csname PYG@tok@gu\endcsname{\let\PYG@bf=\textbf\def\PYG@tc##1{\textcolor[rgb]{0.50,0.00,0.50}{##1}}}
\expandafter\def\csname PYG@tok@gt\endcsname{\def\PYG@tc##1{\textcolor[rgb]{0.00,0.27,0.87}{##1}}}
\expandafter\def\csname PYG@tok@gs\endcsname{\let\PYG@bf=\textbf}
\expandafter\def\csname PYG@tok@gr\endcsname{\def\PYG@tc##1{\textcolor[rgb]{1.00,0.00,0.00}{##1}}}
\expandafter\def\csname PYG@tok@cm\endcsname{\let\PYG@it=\textit\def\PYG@tc##1{\textcolor[rgb]{0.25,0.50,0.50}{##1}}}
\expandafter\def\csname PYG@tok@vg\endcsname{\def\PYG@tc##1{\textcolor[rgb]{0.10,0.09,0.49}{##1}}}
\expandafter\def\csname PYG@tok@vi\endcsname{\def\PYG@tc##1{\textcolor[rgb]{0.10,0.09,0.49}{##1}}}
\expandafter\def\csname PYG@tok@mh\endcsname{\def\PYG@tc##1{\textcolor[rgb]{0.40,0.40,0.40}{##1}}}
\expandafter\def\csname PYG@tok@cs\endcsname{\let\PYG@it=\textit\def\PYG@tc##1{\textcolor[rgb]{0.25,0.50,0.50}{##1}}}
\expandafter\def\csname PYG@tok@ge\endcsname{\let\PYG@it=\textit}
\expandafter\def\csname PYG@tok@vc\endcsname{\def\PYG@tc##1{\textcolor[rgb]{0.10,0.09,0.49}{##1}}}
\expandafter\def\csname PYG@tok@il\endcsname{\def\PYG@tc##1{\textcolor[rgb]{0.40,0.40,0.40}{##1}}}
\expandafter\def\csname PYG@tok@go\endcsname{\def\PYG@tc##1{\textcolor[rgb]{0.53,0.53,0.53}{##1}}}
\expandafter\def\csname PYG@tok@cp\endcsname{\def\PYG@tc##1{\textcolor[rgb]{0.74,0.48,0.00}{##1}}}
\expandafter\def\csname PYG@tok@gi\endcsname{\def\PYG@tc##1{\textcolor[rgb]{0.00,0.63,0.00}{##1}}}
\expandafter\def\csname PYG@tok@gh\endcsname{\let\PYG@bf=\textbf\def\PYG@tc##1{\textcolor[rgb]{0.00,0.00,0.50}{##1}}}
\expandafter\def\csname PYG@tok@ni\endcsname{\let\PYG@bf=\textbf\def\PYG@tc##1{\textcolor[rgb]{0.60,0.60,0.60}{##1}}}
\expandafter\def\csname PYG@tok@nl\endcsname{\def\PYG@tc##1{\textcolor[rgb]{0.63,0.63,0.00}{##1}}}
\expandafter\def\csname PYG@tok@nn\endcsname{\let\PYG@bf=\textbf\def\PYG@tc##1{\textcolor[rgb]{0.00,0.00,1.00}{##1}}}
\expandafter\def\csname PYG@tok@no\endcsname{\def\PYG@tc##1{\textcolor[rgb]{0.53,0.00,0.00}{##1}}}
\expandafter\def\csname PYG@tok@na\endcsname{\def\PYG@tc##1{\textcolor[rgb]{0.49,0.56,0.16}{##1}}}
\expandafter\def\csname PYG@tok@nb\endcsname{\def\PYG@tc##1{\textcolor[rgb]{0.00,0.50,0.00}{##1}}}
\expandafter\def\csname PYG@tok@nc\endcsname{\let\PYG@bf=\textbf\def\PYG@tc##1{\textcolor[rgb]{0.00,0.00,1.00}{##1}}}
\expandafter\def\csname PYG@tok@nd\endcsname{\def\PYG@tc##1{\textcolor[rgb]{0.67,0.13,1.00}{##1}}}
\expandafter\def\csname PYG@tok@ne\endcsname{\let\PYG@bf=\textbf\def\PYG@tc##1{\textcolor[rgb]{0.82,0.25,0.23}{##1}}}
\expandafter\def\csname PYG@tok@nf\endcsname{\def\PYG@tc##1{\textcolor[rgb]{0.00,0.00,1.00}{##1}}}
\expandafter\def\csname PYG@tok@si\endcsname{\let\PYG@bf=\textbf\def\PYG@tc##1{\textcolor[rgb]{0.73,0.40,0.53}{##1}}}
\expandafter\def\csname PYG@tok@s2\endcsname{\def\PYG@tc##1{\textcolor[rgb]{0.73,0.13,0.13}{##1}}}
\expandafter\def\csname PYG@tok@nt\endcsname{\let\PYG@bf=\textbf\def\PYG@tc##1{\textcolor[rgb]{0.00,0.50,0.00}{##1}}}
\expandafter\def\csname PYG@tok@nv\endcsname{\def\PYG@tc##1{\textcolor[rgb]{0.10,0.09,0.49}{##1}}}
\expandafter\def\csname PYG@tok@s1\endcsname{\def\PYG@tc##1{\textcolor[rgb]{0.73,0.13,0.13}{##1}}}
\expandafter\def\csname PYG@tok@ch\endcsname{\let\PYG@it=\textit\def\PYG@tc##1{\textcolor[rgb]{0.25,0.50,0.50}{##1}}}
\expandafter\def\csname PYG@tok@m\endcsname{\def\PYG@tc##1{\textcolor[rgb]{0.40,0.40,0.40}{##1}}}
\expandafter\def\csname PYG@tok@gp\endcsname{\let\PYG@bf=\textbf\def\PYG@tc##1{\textcolor[rgb]{0.00,0.00,0.50}{##1}}}
\expandafter\def\csname PYG@tok@sh\endcsname{\def\PYG@tc##1{\textcolor[rgb]{0.73,0.13,0.13}{##1}}}
\expandafter\def\csname PYG@tok@ow\endcsname{\let\PYG@bf=\textbf\def\PYG@tc##1{\textcolor[rgb]{0.67,0.13,1.00}{##1}}}
\expandafter\def\csname PYG@tok@sx\endcsname{\def\PYG@tc##1{\textcolor[rgb]{0.00,0.50,0.00}{##1}}}
\expandafter\def\csname PYG@tok@bp\endcsname{\def\PYG@tc##1{\textcolor[rgb]{0.00,0.50,0.00}{##1}}}
\expandafter\def\csname PYG@tok@c1\endcsname{\let\PYG@it=\textit\def\PYG@tc##1{\textcolor[rgb]{0.25,0.50,0.50}{##1}}}
\expandafter\def\csname PYG@tok@o\endcsname{\def\PYG@tc##1{\textcolor[rgb]{0.40,0.40,0.40}{##1}}}
\expandafter\def\csname PYG@tok@kc\endcsname{\let\PYG@bf=\textbf\def\PYG@tc##1{\textcolor[rgb]{0.00,0.50,0.00}{##1}}}
\expandafter\def\csname PYG@tok@c\endcsname{\let\PYG@it=\textit\def\PYG@tc##1{\textcolor[rgb]{0.25,0.50,0.50}{##1}}}
\expandafter\def\csname PYG@tok@mf\endcsname{\def\PYG@tc##1{\textcolor[rgb]{0.40,0.40,0.40}{##1}}}
\expandafter\def\csname PYG@tok@err\endcsname{\def\PYG@bc##1{\setlength{\fboxsep}{0pt}\fcolorbox[rgb]{1.00,0.00,0.00}{1,1,1}{\strut ##1}}}
\expandafter\def\csname PYG@tok@mb\endcsname{\def\PYG@tc##1{\textcolor[rgb]{0.40,0.40,0.40}{##1}}}
\expandafter\def\csname PYG@tok@ss\endcsname{\def\PYG@tc##1{\textcolor[rgb]{0.10,0.09,0.49}{##1}}}
\expandafter\def\csname PYG@tok@sr\endcsname{\def\PYG@tc##1{\textcolor[rgb]{0.73,0.40,0.53}{##1}}}
\expandafter\def\csname PYG@tok@mo\endcsname{\def\PYG@tc##1{\textcolor[rgb]{0.40,0.40,0.40}{##1}}}
\expandafter\def\csname PYG@tok@kd\endcsname{\let\PYG@bf=\textbf\def\PYG@tc##1{\textcolor[rgb]{0.00,0.50,0.00}{##1}}}
\expandafter\def\csname PYG@tok@mi\endcsname{\def\PYG@tc##1{\textcolor[rgb]{0.40,0.40,0.40}{##1}}}
\expandafter\def\csname PYG@tok@kn\endcsname{\let\PYG@bf=\textbf\def\PYG@tc##1{\textcolor[rgb]{0.00,0.50,0.00}{##1}}}
\expandafter\def\csname PYG@tok@cpf\endcsname{\let\PYG@it=\textit\def\PYG@tc##1{\textcolor[rgb]{0.25,0.50,0.50}{##1}}}
\expandafter\def\csname PYG@tok@kr\endcsname{\let\PYG@bf=\textbf\def\PYG@tc##1{\textcolor[rgb]{0.00,0.50,0.00}{##1}}}
\expandafter\def\csname PYG@tok@s\endcsname{\def\PYG@tc##1{\textcolor[rgb]{0.73,0.13,0.13}{##1}}}
\expandafter\def\csname PYG@tok@kp\endcsname{\def\PYG@tc##1{\textcolor[rgb]{0.00,0.50,0.00}{##1}}}
\expandafter\def\csname PYG@tok@w\endcsname{\def\PYG@tc##1{\textcolor[rgb]{0.73,0.73,0.73}{##1}}}
\expandafter\def\csname PYG@tok@kt\endcsname{\def\PYG@tc##1{\textcolor[rgb]{0.69,0.00,0.25}{##1}}}
\expandafter\def\csname PYG@tok@sc\endcsname{\def\PYG@tc##1{\textcolor[rgb]{0.73,0.13,0.13}{##1}}}
\expandafter\def\csname PYG@tok@sb\endcsname{\def\PYG@tc##1{\textcolor[rgb]{0.73,0.13,0.13}{##1}}}
\expandafter\def\csname PYG@tok@k\endcsname{\let\PYG@bf=\textbf\def\PYG@tc##1{\textcolor[rgb]{0.00,0.50,0.00}{##1}}}
\expandafter\def\csname PYG@tok@se\endcsname{\let\PYG@bf=\textbf\def\PYG@tc##1{\textcolor[rgb]{0.73,0.40,0.13}{##1}}}
\expandafter\def\csname PYG@tok@sd\endcsname{\let\PYG@it=\textit\def\PYG@tc##1{\textcolor[rgb]{0.73,0.13,0.13}{##1}}}

\def\PYGZbs{\char`\\}
\def\PYGZus{\char`\_}

\def\PYGZca{\char`\^}

\def\PYGZhy{\char`\-}

\makeatother

\usepackage{url}  
\usepackage{graphicx}  
\usepackage{amsmath}
\usepackage{listings}
\usepackage{subfig}
\usepackage{algorithm,txfonts}
\usepackage{algpseudocode}
\newtheorem{theorem}{Definition}

\begin{document}

\title{Learning Probabilistic Logic Programs in Continuous Domains}
\author{Stefanie Speichert  \\
University of Edinburgh \And Vaishak Belle \\ 
University of Edinburgh \& \\ Alan Turing Institute \\}
\maketitle

\begin{abstract}
The field of statistical relational learning aims at unifying logic and probability to reason and learn from data. Perhaps the most successful  paradigm in the field is probabilistic logic programming: the enabling of stochastic primitives in logic programming, which is now increasingly seen to provide a declarative background to complex machine learning applications. While many systems offer inference capabilities, the more significant challenge is that of learning meaningful and interpretable symbolic representations from data. In that regard, inductive logic programming and related techniques have paved much of the way for the last few decades.

Unfortunately, a major limitation of this exciting landscape is that much of the work is limited to finite-domain discrete probability distributions. Recently, a handful of systems have been extended to represent and perform inference with continuous distributions. The problem, of course, is that classical solutions for inference are either restricted to well-known parametric families (e.g., Gaussians) or resort to sampling strategies that provide correct answers only in the limit. When it comes to learning, moreover, inducing representations remains entirely open, other than ``data-fitting" solutions that force-fit points to aforementioned parametric families.

In this paper, we take the first steps towards inducing probabilistic logic programs for continuous and mixed discrete-continuous data, without being pigeon-holed to a fixed set of distribution families. Our key insight is to leverage techniques from piecewise polynomial function approximation theory, yielding a principled way to learn and compositionally construct density functions. 
We test the framework and discuss the learned representations.
\end{abstract}

\section{Introduction} 
\label{sec:introduction}

The field of statistical relational learning aims at unifying logic and probability to reason and learn from relational data. Logic provides a means to codify high-level dependencies between individuals, enabling \emph{descriptive clarity} in the knowledge representation system. Perhaps the most successful  paradigm in the field is \emph{probabilistic logic programming} (PLP): the enabling of stochastic primitives in logic programming. 
Programmatic abstractions further enable {\it modularity} and {\it compositionality}, and are now increasingly seen as providing a much needed declarative interface for complex machine learning applications \cite{de2015probabilistic}.

While a great deal of attention has been paid to the semantics and inference computations of such programming languages \cite{DBLP:journals/tplp/BaralGR09,de2015probabilistic,DBLP:conf/ijcai/MilchMRSOK05}, it is the learning of representations that is deeply  challenging. \emph{Parameter learning} attempts to obtain the probabilities of atoms from  observational traces (e.g., number of heads observed in a sequence of coin tosses) \cite{GutmannECML11,bellodi2011learning}. The significantly harder  problem is that of \emph{structure learning}: for example, learning (deterministic) rules -- essentially, logic programs --  from data. The influential work on \emph{inductive logic programming} and first-order rule learning \cite{muggleton1995inverse,quinlan1990learning} is a major step in this direction. 
Viewed from the perspective of program synthesis \cite{gulwani2010dimensions}, it is worthwhile to remark that  the learning objectives, the synthesis process and the final outcome are all expressed in the same language \cite{deville1994logic}. 
Rule learning is now widely used in \emph{natural language processing} (NLP) applications for  Web data \cite{schoenmackers2010learning}, among others. Naturally, the hardest variant here is to  additionally learn probabilistic atoms together with these rules (e.g., \cite{raghavan2012learning}), so as to yield a probabilistic logic program. This is done by first learning deterministic rules, and then the weights are determined in a second step using parameter estimation techniques.

Unfortunately, a major limitation of this exciting landscape is that much of the work is limited to finite-domain discrete probability distributions. This is a very serious limitation because for many forms of data -- including time-series data, such as temperature observations, trajectories of moving objects, and financial data -- continuous representations are the most natural and compact. Disciplines from robotics to social sciences and biology formulate their findings using continuous, and mixed discrete-continuous probability distributions. Admittedly, there has been some effort in representing and inferring with continuous distributions in a logic programming context (e.g., \cite{gutmann2010extending,307537,nitti2016learning}). 

But inference is difficult; it requires one to either restrict to well-known parametric families (e.g., Gaussians) for efficiency or choose proposal distributions carefully, and then resort to sampling strategies that provide correct answers only in the limit. When it comes to learning, however, inducing representations remains almost entirely open. Conventional ``data-fitting" solutions, for example, force-fit points to parametric families \cite{murphy2012machine}. In fact, to the best of our knowledge, the only approach that discusses the learning of probabilistic logic programs in continuous settings is that of \cite{nitti2016learning}, but this assumes that base distributions (i.e., probabilistic atoms) are Gaussian, and then develops a decision-tree learner for inducing rules from observational traces of these atoms.

In this paper, we take  steps to fill this gap. 
We make three key contributions. First, we address the problem of learning continuous probabilistic atoms but without being pigeon-holed to a fixed set of distribution families; indeed, the shape of the underlying distribution can be arbitrarily complex by appealing to piecewise polynomial approximations \cite{shenoy2011inference}. Efficient integration is possible for that representation \cite{baldoni2014user}.
Second, (discrete) predicates denoting the pieces from that base distribution are then used to learn deterministic rules, yielding dependencies between sub-spaces of a mixed discrete-continuous probability space. What is particularly attractive about this strategy is that the rule learning can be performed using any standard discrete learner. Third, by interfacing a symbolic integration algorithm with a discrete PLP system, we obtain a modular approach for inferring with PLPs in continuous domains. While much of the underlying machinery is agnostic about the PLP language, we develop and implement our approach on ProbLog and its continuous extension \cite{gutmann2010extending} for the sake of concreteness. In a subsequent section, we report on empirical evaluations and discuss the learned representations. To the best of our knowledge, this is the first attempt to induce probabilistic logic programs over continuous features without making any assumptions about the underlying true density, and we hope it will make probabilistic knowledge representation systems more applicable for big uncertain data.

\section{Related Work}

Inference and learning in probabilistic systems are fundamental problems within AI, to which our work here contributes. We begin by remarking that there is an important distinction to be made between relational graphical models 
\cite{richardson2006markov} and the inductive logic programming machinery that we use here \cite{raedt2016statistical}. A comprehensive discussion on the subtleties would be out of scope, and orthogonal to the main thrust of the paper.

The majority of the literature focuses on inference and learning  with  discrete random variables, e.g., \cite{chavira2008probabilistic,getoor2001learning}. 

Nonetheless, learning relational features from data is very popular in NLP and related areas \cite{raghavan2012learning,schoenmackers2010learning}. Rule learning has been studied in many forms, e.g.,  \cite{dvzeroski1993using,landwehr2005nfoil}. 

Approaches such as \cite{zettlemoyer2005learning} have further applied rule learning to complex applications such as automated planning. 

Treating continuous and hybrid data in such contexts, however, is rare. Part of the problem is that inference in mixed discrete-continuous distributions is already very challenging, and learning typically makes use of inference computations to guess good models. Existing inference schemes for hybrid data are either approximate, e.g., \cite{murphy1999variational}, or make restrictive assumptions about the distribution family (e.g., Gaussian potentials \cite{lauritzen2001stable}). Structure learning schemes, consequently, inherit these limitations, e.g., \cite{HGC95}.

In the PLP community, \cite{nitti2016learning} discussed above, learn rules by assuming Gaussian base atoms. 

The restrictive nature of parametric families has led to intense activity on piecewise polynomial constructions, e.g., \cite{shenoy2011inference,sanner2012symbolic,belle2015probabilistic}. 
 On the one hand, this representation is general, in that it can be made arbitrarily close to non-polynomial density functions (such as normal and log-normal), by increasing the degree of polynomials and the granularity of the piecewise composition (e.g., \cite{shenoy2011inference,lopez2014learning}). On the other, under mild conditions, it supports efficient integration (e.g., \cite{baldoni2014user,albarghouthi2017quantifying}). 
Recently, weighted model integration (WMI) was proposed as a computational abstraction for computing probabilities with continuous and mixed discrete-continuous distributions \cite{belle2015probabilistic,RupakSMT,albarghouthi2017quantifying} based on piecewise polynomials. It generalises weighted model counting that is defined over propositional formulas for finite-domain discrete probability distributions, which is the backbone of ProbLog's inference engine \cite{FierensUAI11}. 

\section{Preliminaries}

\subsection{Logic Programming}
First, we recall the basics of Prolog and logic programming. A term $t$ denotes either a constant, a variable or a functor. 

An \textit{atom} $p(t_1,..,t_n)$ is obtained by applying terms $t_1, \ldots, t_n$ to 
the $n$-ary relation $p$.
 
  An atom is called \textit{ground} if all its terms are constants. 

A \textit{literal} denotes an atom or its negation. A \textit{clause} is a disjunction of literals. 

If a clause only contains one non-negated atom, the clause is called \emph{definite}. Such clauses are written $h \leftarrow b_1,...,b_n$, where $h$ is the called the \emph{head}, and the rest the body, all of which are atoms. A Prolog program is a set of definite clauses. 

Variables $V_i$ are mapped to terms $t_i$ through \textit{substitution}: 
\[ \theta = \{V_1/t_1,...,V_n/t_n\}
\]
A \textit{grounding substitution} for an atom $a\theta$ maps all its logical variables to constants.

\subsection{ProbLog}
ProbLog is a probabilistic extension of Prolog \cite{de2007problog}. In addition to the set of definite clauses $D$, described above, it allows the specification of probabilistic facts, background knowledge and evidence. Facts $c_i$ are given probabilities $p_i$ to build a set of \textit{probabilistic facts} that form the basis of ProbLog programs: $T=\{p_1::c_1,\ldots,p_n::c_n\}$. 
The probabilities model how likely it is that a grounding instance $c_i\theta$ is true. Given a set $L_T=\{c_1\theta_{1},\ldots, c_n\theta_{m}\}$ of such instances ProbLog defines a probability distribution over subsets of  facts $L \subseteq L_T$ as:
\[P(L\mid T)=\prod_{c_i\theta_j \in L} p_i \prod_{c_i\theta_j \in L_T \setminus L} (1-p_i)
\]
The success probability of a query $q$ is then defined as:
\[ P_s(q\mid T)= \sum_{L \subseteq L_T\colon L \cup D \models q}P(L|T)
\]
Exact inference for a ProbLog program is performed by converting the program space $L$ to a set of weighted Boolean formulas $\phi$, and computing the sum of the weights of interpretations \cite{FierensUAI11}.

\subsection{ProbFOIL}
Structure learning was recently discussed using the approach of ProbFOIL, a probabilistic extension of classical FOIL learners \cite{de2015inducing}. It even allows for noisy examples.  The problem is defined as:

\begin{theorem}
\cite{de2015inducing} 
Given:
\begin{enumerate}
\item A set of examples $E$, consisting of pairs $(x_i,p_i)$ where $x_i$ is a ground fact for the unknown target predicate t and $p_i$ is the target probability.
\item A background theory $B$ containing information about the examples in the form of a ProbLog program;
\item A loss function $\mbox{loss}(H,B,E)$, measuring the loss of a hypothesis (set of clauses) $H$ w.r.t $B$ and $E$;
\item A space of possible clauses $L_h$ specified using a declarative bias;
 \end{enumerate}
Find: A hypothesis $H \subseteq L_H$ such that $H = \mbox{arg min\_H}  \quad loss(H, B, E)$.

\end{theorem}

(We omit the definition of the loss function.) 
For simplicity, in this paper, we use ProbFOIL in a deterministic/noise-free setting (i.e., $p_i=1$).

\subsection{Hybrid ProbLog}
Hybrid ProbLog \cite{gutmann2010extending} extends ProbLog to support continuous densities in \emph{parametric form}, such as Gaussians. In addition to the set of probabilistic facts, hybrid ProbLog consists of probabilistic continuous facts:
\[ F^c=\{(X_1,\phi_1)::f_1^c,...(X_m,\phi_m)::f_m^c\}
\]
Here, $X_i$ a Prolog variable that is bound to a (say)  Gaussian density $\phi_i$ that belongs to the atom $f_i^c$. For example, {\scriptsize 
\begin{Verbatim}[commandchars=\\\{\}]
\PYG{p}{(}\PYG{n+nv}{I}\PYG{p}{,}\PYG{n+nv}{Gaussian}\PYG{p}{(}\PYG{l+m+mi}{90}\PYG{p}{,}\PYG{l+m+mi}{10}\PYG{p}{))} \PYG{l+s+sAtom}{::} \PYG{n+nf}{intelligence}\PYG{p}{(}\PYG{n+nv}{I}\PYG{p}{).}
\end{Verbatim}
}
says that the intelligence (of, say, students in some class) modelled as a numeric value (such as IQ)  is normally distributed around a mean value of 90 and standard deviation 10. Analogously, a Gaussian mixture model can be expressed using: {\scriptsize 

\begin{Verbatim}[commandchars=\\\{\}]
\PYG{l+m+mf}{0.6} \PYG{l+s+sAtom}{::} \PYG{l+s+sAtom}{heads}\PYG{p}{.}
\PYG{p}{(}\PYG{n+nv}{I}\PYG{p}{,}\PYG{n+nv}{Gaussian}\PYG{p}{(}\PYG{l+m+mi}{110}\PYG{p}{,}\PYG{l+m+mi}{10}\PYG{p}{))} \PYG{l+s+sAtom}{::} \PYG{n+nf}{intelligence\PYGZus{}smart}\PYG{p}{(}\PYG{n+nv}{I}\PYG{p}{).}
\PYG{n+nf}{mix}\PYG{p}{(}\PYG{n+nv}{I}\PYG{p}{)} \PYG{p}{:\PYGZhy{}} \PYG{l+s+sAtom}{heads}\PYG{p}{,} \PYG{n+nf}{intelligence}\PYG{p}{(}\PYG{n+nv}{I}\PYG{p}{).}
\PYG{n+nf}{mix}\PYG{p}{(}\PYG{n+nv}{I}\PYG{p}{)} \PYG{p}{:\PYGZhy{}} \PYG{n+nf}{intelligence\PYGZus{}smart}\PYG{p}{(}\PYG{n+nv}{I}\PYG{p}{),} \PYG{l+s+sAtom}{\PYGZbs{}+} \PYG{l+s+sAtom}{heads}\PYG{p}{.}
\end{Verbatim}
}
In order to perform inference and query in this language, 3 predicates are additionally introduced: \cite{gutmann2010extending}:
\begin{itemize}
    \item {\tt below}$(X,c)$: succeeds if X can be grounded to a continuous value and $X < c$ for the constant c. 
    \item {\tt above}$(X,c)$: succeeds if X can be grounded to a continuous value and $X > c$ for the constant c. 
    \item {\tt ininterval}$(X,c_1,c_2)$ succeeds if X can be grounded to a continuous value and $c_1 \leq X \leq c_2$ for two constants $c_1,c_2$. 
\end{itemize}
Queries are build by binding a continuous relation to an interval. For example: {\scriptsize 
\begin{Verbatim}[commandchars=\\\{\}]
\PYG{n+nf}{average} \PYG{o}{:\PYGZhy{}} \PYG{n+nf}{intelligence}\PYG{p}{(}\PYG{n+nv}{I}\PYG{p}{),} \PYG{n+nf}{ininterval}\PYG{p}{(}\PYG{n+nv}{I}\PYG{p}{,}\PYG{l+m+mi}{65}\PYG{p}{,}\PYG{l+m+mi}{85}\PYG{p}{).}
\end{Verbatim}
}
says that {\tt average} is true if the intelligence measure is between 65 and 85. Naturally, given the above distribution, one could ask: {\scriptsize 
\begin{Verbatim}[commandchars=\\\{\}]
\PYG{n+nf}{query}\PYG{p}{(}\PYG{l+s+sAtom}{average}\PYG{p}{).}
\end{Verbatim}
}
This would return the probability for the interval $[65,85]$ as determined by the Gaussian above.

 In general, the success probability of query is an adaptation of the discrete case and takes the form: \[
	P_s(q\mid T) = \sum_{L\subseteq L_T} \sum_{I\in A\colon L\cup L_I \cup D \models q} P(L\mid T) \cdot \delta.
\]
The intuition here is this. The query $q$ clearly determines a subspace of the domain of a random variable (i.e., $[65,85]$), and so imagine $A$ to partition $[-\infty,\infty]$ to intervals, one of which contains $q$. 
Thus, the success probability is defined by summing over the elements of this partition (e.g., $< 65$, $q$ and $> 85$), with the understanding that $L_I$ expresses this interval (by relativising w.r.t. $\tt above, below, ininterval$). For each such interval, we compute the success of $q$ w.r.t. the product of density functions $\delta.$ In practice, the algorithm for computing these probabilities proceeds by realising such a partition, computing integrals for the intervals and then essentially reverting to the discrete  program space \cite{gutmann2010extending}.

\section{Framework}

The aim of our framework is to provide a principled \textit{unsupervised} way to learn hybrid (PLP) programs from data and to support exact inference over univariate as well as multivariate densities. 

\begin{figure}[h]
  \centering
  \subfloat[Approximation  12 intervals] {\includegraphics[width=4cm]{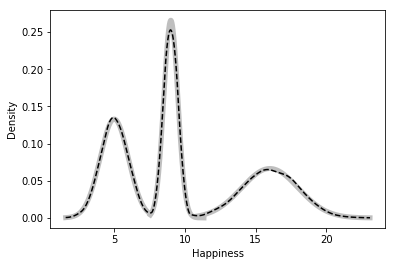}\label{fig:f1}}
  \hfill
  \subfloat[Approximation 2 intervals]{\includegraphics[width=4cm]{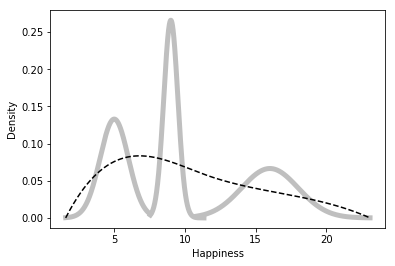}\label{fig:f2}}
  \caption{The fictional happiness attribute}
  \label{fig:comparison}
\end{figure}

For this, to avoid being pigeon-holed to a fixed set of distributions, we appeal to piecewise polynomials that approximate any density arbitrarily close. (Cf. discussion on piecewise polynomial density approximations). 
Furthermore, we show how to learn the optimal underlying piecewise structure  and how to leverage that piecewise structure to learn rules to  yield a granular hybrid program. An example of a good and a bad approximation can be found in Figure \ref{fig:comparison}.

While much of the underlying learning machinery is agnostic about the PLP language, we develop our framework on ProbLog for the sake of concreteness. 

Mainly, we utilise and extend the syntax of Hybrid ProbLog and show how to use ProbLog's discrete rule learner ProbFOIL to learn hybrid programs. We reiterate that this is an important feature.

While some prior accounts have taken steps towards extending a language to continuous domains (w.r.t. restricted, often parametric, families), none show how a standard discrete structure learner suffices. The ability to learn arbitrary PDFs together with  effective inference \cite{baldoni2011integrate}  makes the framework, in our view, novel and powerful. 

\section{Inference with Piecewise Polynomials}
This section discusses our first contribution. To prepare for the learning of arbitrary PDFs, we need to 
revisit hybrid ProbLog. As discussed, we propose to model  density functions as piecewise polynomials (PP). On the one hand,  piecewise polynomial representations can be made arbitrarily close to any distribution  \cite{shenoy2011inference}. On the other hand, these representations are amenable to effective integration \cite{baldoni2011integrate}, scaling to hundreds of variables, e.g., \cite{belle2015probabilistic}. Our first contribution discusses how to: (a) model PP densities in Hybrid ProbLog with a slightly revised syntax, but without affecting the semantic devices; and (b) leverage symbolic integration techniques to compute success probabilities.

Let us begin with the syntax, and expected behaviour of queries. 
\begin{theorem} A piecewise function over a real-valued variable $x$ is defined over $l$ pieces  as:
\[ \vec{pp}(X)= \begin{cases} 
      0 & x < \textit{cp}_0 \\
      {pp}_1(x) & \textit{cp}_{0}\leq x\leq \textit{cp}_1 \\
      ...\\
      {pp}_l(x) & \textit{cp}_{l_1} \leq x \leq \textit{cp}_l \\
      0 &  x > \textit{cp}_l
   \end{cases}
\]
where the intervals (expressed using \emph{cutpoints}) are mutually exclusive, and 
each ${pp}_i(x)$ is a polynomial with the same maximum polynomial order $k$ of the form: 
\[{pp}_i(x)= b_0^i + b_1^i*x +...+ b_k^i*x^k.\]
In order for $\vec{pp}(x)$ to form a valid density,  $\sum_{i=1}^l\int_{{cp}_{i-1}}^{{cp}_i} {pp}_i(x) dx =1$.  
	
\end{theorem}
Figure \ref{fig:comparison} demonstrates the importance of choosing the right parameters for each interval in order to approximate the function well.

\newcommand{\btw}{\textit{between}}
\newcommand{\inl}{\textit{intelligence}}

When it comes to the program itself, then, for every continuous random variable, a new relation is added corresponding to each piece, defined using the original continuous attribute and further relativisation using the predicates $\texttt{above, below}$ and $ \texttt{ininterval}$. For example, reconsider the intelligence random variable from above, and suppose we would like to approximate the Gaussian using a 5-piece piecewise polynomial density function. 

The program might include sentences such as: 
{
\scriptsize 
\begin{Verbatim}[commandchars=\\\{\}]
\PYG{o}{\PYGZhy{}}\PYG{l+m+mf}{0.024719432823743857} \PYG{o}{+} \PYG{l+m+mf}{0.0005171566890546171} \PYG{n+nv}{I}  \PYG{l+s+sAtom}{::} \PYG{n+nf}{int\PYGZus{}low}\PYG{p}{(}\PYG{n+nv}{I}\PYG{p}{).}
\PYG{n+nf}{int\PYGZus{}low}\PYG{p}{(}\PYG{n+nv}{I}\PYG{p}{)} \PYG{p}{:\PYGZhy{}} \PYG{n+nf}{intelligence}\PYG{p}{(}\PYG{n+nv}{I}\PYG{p}{),} \PYG{n+nf}{below}\PYG{p}{(}\PYG{n+nv}{I}\PYG{p}{,}\PYG{l+m+mi}{70}\PYG{p}{).}
\PYG{n+nf}{int\PYGZus{}mid}\PYG{p}{(}\PYG{n+nv}{I}\PYG{p}{)} \PYG{p}{:\PYGZhy{}} \PYG{n+nf}{intelligence}\PYG{p}{(}\PYG{n+nv}{I}\PYG{p}{),} \PYG{n+nf}{ininterval}\PYG{p}{(}\PYG{n+nv}{I}\PYG{p}{,}\PYG{l+m+mi}{70}\PYG{p}{,}\PYG{l+m+mi}{90}\PYG{p}{).}
\end{Verbatim}
}
What is particularly interesting about this reformulated program is that the partitioning of the real space needed for computing success probabilities in Hybrid ProbLog is already in place, and moreover, syntactically it resembles standard ProbLog, except for non-numeric weights on atoms. 
Thus, the success probability of query can be \emph{computed  externally} and returned to (standard) ProbLog to be used in the evaluation of the program. 

To understand how that works, suppose now,  we are interested in the probability of $\tt average$ as before. The framework then splits the query into two relations by using the intervals from the program: 
{\scriptsize \begin{Verbatim}[commandchars=\\\{\}]
\PYG{n+nf}{average1} \PYG{o}{:\PYGZhy{}} \PYG{n+nf}{intelligence}\PYG{p}{(}\PYG{n+nv}{I}\PYG{p}{),}\PYG{n+nf}{ininterval}\PYG{p}{(}\PYG{n+nv}{I}\PYG{p}{,}\PYG{l+m+mi}{65}\PYG{p}{,}\PYG{l+m+mi}{70}\PYG{p}{).}
\PYG{n+nf}{average2} \PYG{o}{:\PYGZhy{}} \PYG{n+nf}{intelligence}\PYG{p}{(}\PYG{n+nv}{I}\PYG{p}{),}\PYG{n+nf}{ininterval}\PYG{p}{(}\PYG{n+nv}{I}\PYG{p}{,}\PYG{l+m+mi}{70}\PYG{p}{,}\PYG{l+m+mi}{85}\PYG{p}{).}
\end{Verbatim}
}

The probability of $\tt average1$ can be computed using the PP density for $\tt int\_low(I)$ using: $\int_{65}^{70} -0.024719432823743857 + 0.0005171566890546171 x  dx$, and that of $\tt average2$ can be computed using the PP density for $\tt int\_mid(I)$ (omitted).  
It then follows that the success probability of $\tt average$ is the sum of $\tt average1$ and $\tt average2.$

In general, as the density is defined as a {piecewise} polynomial $\vec{pp}(x)$ with $l$ pieces and cutpoints $[\textit{cp}_{i-1},\textit{cp}_i]$, $i=1,...,l$ the total area is obtained as follows: 
\[\int_{\textit{cp}_0}^{\textit{cp}_1} \textit{pp}_1(x) \textit{dx} + ... + \int_{\textit{cp}_{l-1}}^{\textit{cp}_l} \textit{pp}_l(x) \textit{dx}
\]
Note that although $x$ is a real-valued variable and so the domain is defined on $[-\infty,\infty]$, by definition it is 0 outside of $[\textit{cp}_0,\textit{cp}_l]$. 

Therefore, the probability of a query $x\in [a,b]$ is computed as:
\[P(x \in [a,b]) =\int_{a}^{\textit{cp}_i} \textit{pp}_i(x) \textit{dx} + ... + \int_{\textit{cp}_{j-1}}^{b} \textit{pp}_j(x) \textit{dx}.\] 
Negated atoms, which are then equivalent to $x\not\in [a,b]$ are obtained by applying $P(x\not\in[a,b]) = 1 - P(x\in [a,b])$, because the PPs specified in the program are assumed to define a valid density. (In the following section, we discuss how to learn valid probability densities.) 

To reiterate, the attractiveness of this approach is that it allows us to handle continuous computations externally and return success probabilities of (final or intermediate) query atoms to (classical) ProbLog. It seems to us that such a scheme is very much in the spirit of the original Hybrid ProbLog proposal \cite{gutmann2010extending}, but achieves a more direct path from program transformation to  inference computation. 

Perhaps the most serious limitation of \cite{gutmann2010extending} is that the computational machinery is restricted to univariate distributions: logically speaking, unary relations. We show later that this limitation does not apply to our inference engine.

\section{Learning Weighted Atoms}

This section discusses our second contribution. 
We  present a fully \textit{unsupervised approach} to jointly learn intervals and their piecewise polynomial densities from data. The goal of this learner is to divide the attribute into $l$ optimal pieces such that the PP density approximates the data well without any knowledge of the true density. This is achieved in two steps. First, the attribute is divided into $l$ pieces given a criterion. Then, the PP weights for the intervals are calculated. 

How well PPs estimate the data is determined by a noise reducing scoring function -- the so-called Bayesian Information Criterion (BIC) score \cite{kass1995reference} -- so as to avoid overfitting. The algorithm chooses the best discretisation and PP approximation given a set of discretisation criteria, numbers of pieces and polynomial orders. 

\subsection{Discretisation}

In our context, discretisation refers to dividing the range of a continuous variable into $l$ mutually exclusive intervals according to some criteria. Suppose $X$ is a real-valued variable -- logically, think of the argument in  $\tt intelligence(X)$ -- and suppose $x_{min},\ldots,x_{max}$ are the data points we observe. Discretisation  yields $I = \{ [\textit{cp}_0, \textit{cp}_1],...,[\textit{cp}_{l-1}, \textit{cp}_l] \}$. The set of cutpoints $\textit{CP} = \{\textit{cp}_0,...,\textit{cp}_l \}$  determine the interval such that $\textit{cp}_{i-1} < \textit{cp}_i$, $i\in \{ 1,...,l\}$, $\textit{cp}_0 = x_{min}$ and $\textit{cp}_l=x_{max}$. The discretisation step, therefore, defines intervals over the domain $\Omega=[x_{min},x_{max}]$.

To find the best set of intervals, ideally, the whole space of possible cutpoints should be searched exhaustively.

However, as the attribute is continuous, there exist infinitely many possibilities. To restrict the search as much as possible we aim to find simple discretisation  schemes that fulfil two criteria 
to lay the groundwork for the PP learning:

\begin{enumerate}
\item Cutpoints should span the entire set of data points (i.e., attribute instances). Discretisations that only focus on parts of the attribute's values will never be able to model the  distribution reasonably; cf Figure \ref{fig:bad}.  

\item The discretisation scheme should make use of some density statistics that can be determined in an unsupervised manner. 

\end{enumerate}

Consequently, we chose to use equal-width and equal-frequency discretisation \cite{dougherty1995supervised}. (None of our algorithms hinge on the use of these or other discretisation schemes.)  Both methods are directly comparable as they are regulated by the same parameter $l$ that determines the number of bins.

Equal-width binning divides an attribute into equally wide intervals where the width is calculated as $ \textit{wi}_{ew} = (x_{max} -x_{min}) / l$ and each cutpoint $\textit{cp}_i$ , $i\in \{1,...,l\}$ is defined as: $\textit{cp}_i= \textit{wi}_{ew} * i$. This ensures that all data points are taken into account. It is, however, sensitive towards sparse data as sometimes only a few datapoints will yield an interval and, therefore, lead to inaccurate density estimation on an unseen test set.

\renewcommand{\mbox}{\textit}

In contrast, equal-frequency binning ensures that a (nearly) equal number of elements can be found in each interval: $\mbox{wi}_{ef} = \lfloor |X| /l \rfloor$. The cutpoints determine which index to cut:  $\mbox{cp}_i$ is the $m$th-element of $X$, where $m=[\mbox{wi}_{ef}*i]$ and it is assumed that $X$ is ordered with $x_i \leq x_{i+1}$ for all $i$. While this method provides a more robust division, it is often not very accurate in the tails. 

A longer discussion on binning including a comparison of the two methods can be found in the evaluations section. 

\subsection{Learning Weights}
Once an interval sequence is determined, its piecewise polynomial density can be obtained. 
 We utilise Basis splines \cite{Speichert:Thesis:2017} as a non-parametric density estimator. In contrast to methods such as Taylor series expansions \cite{shenoy2011inference}, where the true density needs to be known  or where samples from the true PDF are used for Lagrange interpolation \cite{shenoy2012two}, this method does not need any prior knowledge about the true underlying PDF. 
In addition, the splines can be modified to accommodate any form of discretisation,  which allows for a clean separation of weight and structure learning. 

Basis splines form a basis in the piecewise polynomial space. By considering linear combinations of splines,  polynomials with a variety of different properties can be obtained. The combination is influenced by a set of mixing coefficients. By imposing constraints on the coefficients, the generated polynomials are of low ($\leq10$) order and are guaranteed to form a valid density. They also possess a number of desirable properties such as being closed under addition, subtraction and multiplication and, therefore, integration and combination. \footnote{A detailed description of the polynomial weight generation will be made available in an extended version of the paper.}

As discussed, the key criterion to determine the best polynomial representation is the BIC score - a penalised log-likelihood scoring function. The score penalises involved (spline) configurations in order to keep the model from overfitting. 

An overview of the algorithm, and how it fits in with the broader picture is given in  Algorithm 1. 
\begin{algorithm}[t]
  \caption{The general loop structure to generate PP representations for a real-valued variable $X$. The parameters \emph{maxSize} and \emph{maxOrder} are set by the user. Usual values are marked as defaults.}\label{euclid}
 {\footnotesize 
  \begin{algorithmic}[]
    \Procedure{\textsc{buildPPstructure}}{$X,\textrm{maxSize}=40,\textrm{maxOrder}=8$}
    \State $X \gets$ sort($X$,ascending)
    \State $\textrm{bestBIC} \gets - \infty$
    \State $\textrm{bestPolyStructure} \gets \textsc{NULL}$
    \For{$l \in~ (2,\textrm{maxSize}):$}
    	\For{$d \in~$ (``eq-width",``eq-freq")}
        	\State \emph{CP} $\gets$ {discretise}($X,d$)
            \For{$k$ in (1,...,\textrm{maxOrder})}
            	\State $\textrm{curRepr} \gets \textrm{calcPP}(\textit{CP},k)$
                \State $\textrm{curBIC} \gets \textrm{calcBIC(curRep)}$
                \If{curBIC $>$ bestBIC}
                	\State $\textrm{bestPolyStructure} \gets \textrm{curRep}$
                    \State $\textrm{bestBIC} \gets \textrm{curBIC}$
                \EndIf
            \EndFor
        \EndFor
   	\EndFor
    \State \textbf{return} \textsc{transformIntoProbLog}(bestPolyStructure)
    \EndProcedure
  \end{algorithmic}
  }
  
\end{algorithm}
\subsection{Learning Rules} 

By leveraging relational rule learners, specifically ProbFOIL,  we move beyond simply learning weighted atoms to learning complex dependencies between subspaces in a mixed discrete-continuous setting. The basic idea is to augment the original dataset that uses continuous variables (such as $\tt intelligence(X)$) together with instances of the invented predicates (such as $\tt int\_low(X)$), as determined by the discretisation. Clearly, this can be used with any discrete rule learner. 

At this stage, there are multiple choices for rule learning. In the simplest setting, we  ignore the learned densities and perform rule learning for some target predicate. (That could be repeated for multiple targets.) This is what constitutes the setting for standard first-order inductive rule learning, e.g., \cite{muggleton1995inverse,quinlan1990learning}, where the background theory is specified as a set of ground facts  and each example is a true or false fact for the target predicate. A variant of this assumes the background knowledge is probabilistic, but the examples themselves are deterministic. A final variant additionally assumes noisy examples \cite{de2015inducing,chen2008learning}. 

In this work, we focused on deterministic examples for simplicity, and the previous sections described ways to induce continuous probabilistic facts. The facts could be ignored, but since ProbFOIL does support learning from deterministic examples and probabilistic background knowledge, we discuss how to transform our atoms to be used with ProbFOIL and related learners. 

We consider a  discretisation scheme $F^C \rightarrow F$ for the hybrid relations by calculating   for each clause  $c_j$ and each piecewise polynomial density $\mbox{pp}_j(x)$ over cutpoints $[\mbox{cp}_{j-1},\mbox{cp}_j]$, $j\in \{ 1,...,l\}$: 
\[ p_j= \int_{c_{j-1}}^{c_j} \mbox{pp}_j(x)dx
\]
where $p_j$ is a constant that denotes the probability over the interval $[\mbox{cp}_{j-1},\mbox{cp}_j]$.
The hybrid atom is now transformed into a standard ProbLog atom $p_j :: c_j$. For example, the hybrid atom 
{\scriptsize 
\begin{Verbatim}[commandchars=\\\{\}]
\PYG{o}{\PYGZhy{}}\PYG{l+m+mf}{0.024719432823743857} \PYG{o}{+} \PYG{l+m+mf}{0.0005171566890546171} \PYG{n+nv}{I} \PYG{l+s+sAtom}{::}  \PYG{n+nf}{int\PYGZus{}low}\PYG{p}{(}\PYG{n+nv}{I}\PYG{p}{).}
\PYG{n+nf}{int\PYGZus{}low}\PYG{p}{(}\PYG{n+nv}{I}\PYG{p}{)} \PYG{p}{:\PYGZhy{}} \PYG{n+nf}{intelligence}\PYG{p}{(}\PYG{n+nv}{I}\PYG{p}{),} \PYG{n+nf}{below}\PYG{p}{(}\PYG{n+nv}{I}\PYG{p}{,}\PYG{l+m+mi}{70}\PYG{p}{).}
\end{Verbatim}
}
is transformed to  
{\scriptsize 
\begin{Verbatim}[commandchars=\\\{\}]
\PYG{l+m+mf}{0.12} \PYG{l+s+sAtom}{::} \PYG{n+nf}{int\PYGZus{}low}\PYG{p}{(}\PYG{n+nv}{I}\PYG{p}{).}
\PYG{n+nf}{int\PYGZus{}low}\PYG{p}{(}\PYG{n+nv}{I}\PYG{p}{)} \PYG{p}{:\PYGZhy{}} \PYG{n+nf}{intelligence}\PYG{p}{(}\PYG{n+nv}{I}\PYG{p}{),} \PYG{n+nf}{below}\PYG{p}{(}\PYG{n+nv}{I}\PYG{p}{,}\PYG{l+m+mi}{70}\PYG{p}{).}
\end{Verbatim}
}
since $\int_{-\infty}^{70} pp(x) dx = 0.12$, where $pp(x)$ is the polynomial specified for {\tt int\_low(I)}.  

This transformation is applied to all continuous predicates and evidence such that the discrete rule learner can interpret them.  

\section{Extensions}

\subsection{Supervised Discretisation} 
\label{sub:supervised_discretisation}

So far, the proposed framework has been entirely unsupervised: we build relations and weights based on the BIC score and learn rules for them. However, since the BIC score often learns a large number of relations, rules become unnecessarily complex and long. Rules such as: {\scriptsize 
\begin{Verbatim}[commandchars=\\\{\}]
\PYG{n+nf}{grade\PYGZus{}low}\PYG{p}{(}\PYG{n+nv}{C}\PYG{p}{)} \PYG{p}{:\PYGZhy{}} \PYG{n+nf}{intelligence}\PYG{p}{(}\PYG{n+nv}{I}\PYG{p}{),} \PYG{n+nf}{ininterval}\PYG{p}{(}\PYG{n+nv}{I}\PYG{p}{,}\PYG{l+m+mi}{60}\PYG{p}{,}\PYG{l+m+mi}{70}\PYG{p}{),}
\PYG{n+nf}{ininterval}\PYG{p}{(}\PYG{n+nv}{I}\PYG{p}{,}\PYG{l+m+mi}{70}\PYG{p}{,}\PYG{l+m+mi}{80}\PYG{p}{),} \PYG{p}{...,} \PYG{n+nf}{course}\PYG{p}{(}\PYG{n+nv}{C}\PYG{p}{).}
\end{Verbatim}
} 
demonstrate that rule bodies can be shortened if relations of lesser granularity are learned, e.g. $\tt ininterval(I,60,80) $. Such rules make the program cleaner and smaller. However, we observed that the BIC score sometimes  naturally selects higher bins  and so we attempted to  look for a different metric. In this section, as a variant of our framework, we  reconsider  the unsupervised paradigm in the context of discretisation. 

Incidentally, many supervised discretisation methods have been introduced in the literature. We refer the reader to \cite{garcia2013survey} for a review. The survey identifies a robust discretisation method, the so-called ``Distance" technique \cite{cerquides1997proposal}, as preprocessing step for rule learning. 

The method stems from information theory and utilises a distance measure based on entropy to find the discretisation with the fewest bins that accurately describes the data. We modified that algorithm so as to produce a specified number of bins, to be chosen by the modeller, so as to be directly comparable   to the other binning methods.

Nonetheless, supervised algorithms do favour fewer numbers of bins. We suspected that this would lead to simpler learned rules and, thus, simpler  programs. It does, however, pose a trade-off between choosing the optimal representation according to either the weights (via the BIC score) or rule learning, as enabled by this supervised algorithm. 
The  evaluation section presents a  detailed report on our findings.

\subsection{Multivariate Inference}

The previous sections have shown how to learn programs and perform inference on discrete and continuous univariate  variables (thus, yielding unary relations). In this section, we discuss the extension in our framework for multivariate random variables. We do not consider learning in this setting currently, and focus solely on inference, which means that they   can also appear as background knowledge. 

Multivariate piecewise polynomials for $m$ real-valued  variables  over hyper-cubes are defined as \cite{shenoy2012two}: 
\[f(x_1,...,x_m)= \begin{cases} 
      P_i(x_1,...,x_m) & (x_1,...,x_m) \in A_i, i\in \{1,...,k\} \\
      0 & else \end{cases}\]
where $P_i$ is a multivariate polynomial, and the variable $A_i$ now describes a hyper-cube  rather than an interval over the reals. That is,  $A_i$ can be defined as $a_{ij} \leq x_j \leq b_{ij}$ for $j\in \{1,\ldots,m\}$, and $a_{ij}, b_{ij}$ are constants.  

Relations can be defined as usual: {
\scriptsize
\begin{Verbatim}[commandchars=\\\{\}]
\PYG{p}{(}\PYG{n+nv}{X\PYGZus{}1}\PYG{p}{)}\PYG{l+s+sAtom}{\PYGZca{}}\PYG{l+m+mi}{2} \PYG{o}{+} \PYG{n+nv}{X\PYGZus{}1}\PYG{o}{*}\PYG{n+nv}{X\PYGZus{}m} \PYG{l+s+sAtom}{::} \PYG{n+nf}{p1}\PYG{p}{(}\PYG{n+nv}{X\PYGZus{}1}\PYG{p}{,...,}\PYG{n+nv}{X\PYGZus{}m}\PYG{p}{).}
\PYG{n+nf}{p1}\PYG{p}{(}\PYG{n+nv}{X\PYGZus{}1}\PYG{p}{,...,}\PYG{n+nv}{X\PYGZus{}m}\PYG{p}{)} \PYG{p}{:\PYGZhy{}} \PYG{n+nf}{p}\PYG{p}{(}\PYG{n+nv}{X\PYGZus{}1}\PYG{p}{,...,}\PYG{n+nv}{X\PYGZus{}m}\PYG{p}{),}
\PYG{n+nf}{ininterval}\PYG{p}{(}\PYG{n+nv}{X\PYGZus{}1}\PYG{p}{,}\PYG{l+s+sAtom}{a\PYGZus{}1}\PYG{p}{,}\PYG{l+s+sAtom}{b\PYGZus{}1}\PYG{p}{),} \PYG{p}{...,} \PYG{n+nf}{ininterval}\PYG{p}{(}\PYG{n+nv}{X\PYGZus{}m}\PYG{p}{,}\PYG{l+s+sAtom}{a\PYGZus{}m}\PYG{p}{,}\PYG{l+s+sAtom}{b\PYGZus{}m}\PYG{p}{).}
\end{Verbatim}
}
Probabilities are computed using the following integral: 
\[ P(x_1 \in [a_1,b_1],...,x_m \in [a_m,b_m]) \]
\[
= \int_{a_1}^{b_1} ... \int_{a_m}^{b^m} PP(x_1,...,x_m)  dx_m...dx_1 
\]
In our experimental evaluations section, we discuss a program where such rules are specified as background knowledge. 

\section{Results}

We report on a number of observations regarding our framework on the following datasets:

\begin{itemize}
\item \textbf{Hybrid University data set:} This  dataset \cite{getoor2001learning} models a semester at a university. Professors teach courses (\texttt{teaches(P,C))}, students take them (\texttt{takes(S,C)}). They receive grades and can rate their satisfaction for each course. The hybrid extension \cite{ravkic2015learning} introduces three new  continuous predicates: \texttt{nrhours(C)} -- modelling the number of hours for a course, \texttt{intelligence(S)} --  the intelligence of students, and \texttt{ability(P)} -- a numeric score denoting the ability of a professor to teach. 

\item \textbf{Happiness Dataset:} This dataset (\url{http://worldhappiness.report/}) ranks countries according to their happiness score. It  introduces  six continuous features: GDP per capita (\texttt{economy(C)}), social security (\texttt{family(C)}), life expectancy  (\texttt{health(C)}), personal freedom  (\texttt{freedom(C)}), absence of corruption  (\texttt{trust(C)}) and generosity (\texttt{generosity(C)}).  In all cases, higher  attribute scores imply better conditions  for the country. 
\end{itemize}
We extend our tests to more hybrid datasets from the UCI repository \cite{Dua:2017}. The datasets are taken from  various domains  such as healthcare and marketing and are of different quality. Some, especially `Anneal-U' and `CRX' contain many missing values. Others contain many duplicates per attribute.   We mainly focus on the university and happiness data to report results but nonetheless briefly discuss other datasets and also remark on how varying levels of data quality affect the learner.

\subsection{Learning Representations}

\begin{table}[]
\centering

\begin{tabular}{lllll}
Dataset    & Train & \# Cont & \# Bins & \% EF  \\ \hline
Anneal-U   & 673           & 6 (39)  & 19      & 100    \\
Australian & 517           & 5 (15)  & 7.5     & 83.333 \\
Auto       & 119           & 15 (26) & 8.733   & 73.333 \\
Car        & 294           & 6 (9)   & 9.2     & 60     \\
Cleave     & 222           & 5 (14)  & 7.75    & 75     \\
Crx        & 488           & 6 (16)  & 11.667  & 100    \\
Diabetes   & 576           & 7 (9)   & 9.128   & 62.5   \\
German     & 750           & 3 (21)  & 6       & 33.333 \\
German-org & 750           & 3 (25)  & 6.333   & 66.667 \\
Iris       & 112           & 4 (5)   & 3.75    & 25    
\end{tabular}
\caption{Statistics on UCI datsets and the Polynomial Learning Component. \# Cont details the number of continuous features with the number of all attributes in brackets. The average number of bins that were learned for each attribute is denoted as \# Bins. \% EF denotes the percentage where the algorithm found the equal-frequency discretisation preferable over equal-width.}
\label{table:polystats}
\end{table}
\subsection{On Piecewise Polynomials}
This section discusses observations for the piecewise polynomial representation learning. 

The BIC score determines the model parameters such as the order of polynomials or, during unsupervised discretisation, the discretisation method and the number of bins. Table \ref{table:polystats} lists statistics for each UCI dataset, which we contextualise further below.

\textbf{Q1: Which trends in parameter learning can be observed for $\ldots$} 

	\textbf{Q1.1:  $\ldots$ the order?}
	
	The order of the polynomials in the learned models stayed relatively low in a range from 2 to 5. In fact, only six attributes over all datasets learned an order that was higher than 5 but never higher than 8. This was observed regardless of the size of the dataset or the non-uniformity of the distribution, which is very desirable. Naturally, low order polynomials are computationally simpler to integrate during inference. The few outliers occurred when an attribute contained a high number of missing values.

	 \textbf{Q1.2: $\ldots$ the number of intervals?}

\begin{figure}[!tbp]
  \centering

  \includegraphics[width=8cm]{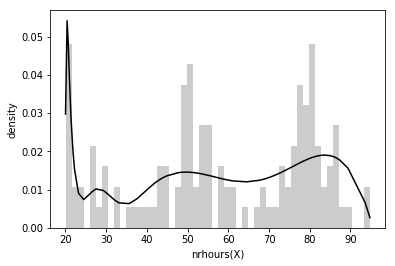}
  \caption{Failure of the binning method for {\texttt nrhours}}
  \label{fig:bad}
\end{figure}

The appropriate number of intervals to approximate a PDF increases with its non-uniformity, as can be seen in Figure \ref{fig:comparison}. The figure depicts two different choices for the number of bins. The method based on two bins approximates the PDF poorly, the method based on 12 bins, however, is very close to the original. Furthermore, to achieve a close approximation, the cutpoints have to span the entire range of attribute values. An illustrative case for this claim was observed when learning the density for the course duration (in terms of numbers of hours) with the university dataset, under the supervised regime. The target attribute is the course difficulty, and lengthier courses were almost always the hard ones. The supervised binning scheme thus failed to recognise that a disproportionate number of data points with the same target value has a course duration of $>45$  hours, and did not create fine enough intervals for those set of values. So the distribution learned for those data points is very poor, as seen in Figure \ref{fig:bad}.

We also observed that the BIC is sensitive to the number of attribute instances. For under 100 data points, it often chose the smallest discretisation and order (cf. Table \ref{table:stats}). This is unfortunate, but other than choosing a different model selection criteria for small datasets, there is not much more to be said here.

\textbf{Q2: How does supervised discretisation compare to the unsupervised regime?} 

Generally speaking, all densities, including ones 
that are outside well-known families and are non-uniform (e.g.,  Figure \ref{fig:f1} and Table\ref{table:stats}) were approximated satisfactorily by unsupervised methods. Approximations by supervised schemes, unsurprisingly, performed worse, in general, as the goal of the discretisation was not an optimal polynomial learner; e.g.,  the previously discussed figure \ref{fig:bad}. When limited to simpler distributions, such as Gaussians, however, the method performed on par or and, in some cases in the happiness dataset, even better than its unsupervised counterpart.

In the unsupervised regime, binning based on equal-frequency was often preferred to equal-width, see Table  \ref{table:polystats} which highlights the percentage of the equal frequency method being used for each dataset. Equal-Width binning, on the other hand, was selected when a small number of bins was able to model the density. This is natural as a small number of bins implies that the underlying distribution is not hard to approximate. If the distributions get more complicated, binning based on equal frequency can model the more complex relations better as it is not dependent on an attribute's spread of values. 

\textbf{Q3: How does the piecewise polynomial approximation fare overall?}

As can be surmised from the above discussions, the PP paradigm is a robust learning strategy for arbitrarily complex PDFs, provided the discretisation is satisfactory. (Both points are illustrated by Figure\ref{fig:f1}, and unsupervised schemes are a safe option to ensure satisfactory discretisation.) 

\subsection{Rule Learning}
{\small  
\begin{table}[]
\centering

\begin{tabular}{@{}lllllllll@{}}

 \texttt{family(C)}  & prec & Neg & Pred& Rules  \\ \hline
Sup: Auto& 0.909 & 0.581 & 2.691&8 \\ 

Sup: 5 Bins &  1.0 & 0.741 & 2.871 &6.6\\ 
Sup: 7 Bins &  1.0 & 0.874 & 2.884 &6.14\\  \hline
Unsup: best& 0.966 & 0.598 & 3.179&10.6\\ 
Unsup: 5 bins & 1.0 & 1.070 & 3.344& 8.4 \\ 
Unsup: 7 bins & 1.0 & 1.562&3.617 & 6.429 \\

\end{tabular}
\caption{Rules learned for all interval instances of the {\tt family} attribute. We compare our unsupervised framework against the extended supervised version. Auto denotes the automatic stopping criterion of the Distance discretisation. The table columns  are: \textbf{prec} = Avg Rule precision over all bins, \textbf{Neg} = Avg number of negations in a rule body, \textbf{Pred} = Avg number of predicates in a rule body, \textbf{Rules} = Avg number of rules in one theory. }
\label{table:stats}

\end{table}

} 
\textbf{Q4: How compact are the learned rules?} \newline 
We admit that compactness is not always a sign of interpretability, and, therefore,  simply report on our empirical observations.
As can be seen in Table \ref{table:stats}, supervised discretisation yields more compact rules than the unsupervised regime: Given the same number of bins, the learned rules are overall shorter, contain fewer negations, and the programs are smaller, too.

A final observation worth reporting is with regards to data preprocessing. A large number of duplicate and missing values seem to lead to lengthy rules in the induced program. Interestingly, regardless of the regime (supervised vs unsupervised), the learned rules for the concerning attributes are almost syntactically identical for each bin size, in clear contrast to table \ref{table:stats}. 
On the one hand, this calls for greater data preprocessing, which is not surprising. Conversely, perhaps the lack of compactness and the syntactic similarity across all regimes could be a diagnostic feature for data preprocessing, an investigation of which we leave for the future. 
\subsection{Example Programs}
Here, we display some sample programs that were compiled after multiple runs of ProbFOIL with different attributes as targets. The polynomial weights have been omitted for readability. (See, for example, the PP density for $\tt int\_low$ from our prior discussions, and the multivariate case below.)

A sample program learned for the university data consists of rules such as:
{\scriptsize 
\begin{Verbatim}[commandchars=\\\{\}]
\PYG{n+nf}{intelligence1} \PYG{o}{:\PYGZhy{}} \PYG{n+nf}{intelligence}\PYG{p}{(}\PYG{n+nv}{I}\PYG{p}{),}\PYG{n+nf}{ininterval}\PYG{p}{(}\PYG{n+nv}{I}\PYG{p}{,}\PYG{l+m+mi}{51}\PYG{p}{,}\PYG{l+m+mi}{60}\PYG{p}{)}
\PYG{n+nf}{intelligence2} \PYG{o}{:\PYGZhy{}} \PYG{n+nf}{intelligence}\PYG{p}{(}\PYG{n+nv}{I}\PYG{p}{),}\PYG{n+nf}{ininterval}\PYG{p}{(}\PYG{n+nv}{I}\PYG{p}{,}\PYG{l+m+mi}{60}\PYG{p}{,}\PYG{l+m+mi}{72}\PYG{p}{)}
\PYG{n+nf}{nrhours2}\PYG{p}{(}\PYG{n+nv}{C}\PYG{p}{)} \PYG{p}{:\PYGZhy{}} \PYG{n+nf}{nrhours}\PYG{p}{(}\PYG{n+nv}{C}\PYG{p}{,}\PYG{n+nv}{N}\PYG{p}{),}\PYG{n+nf}{ininterval}\PYG{p}{(}\PYG{n+nv}{N}\PYG{p}{,}\PYG{l+m+mi}{35}\PYG{p}{,}\PYG{l+m+mi}{50}\PYG{p}{).}

\PYG{n+nf}{satisfaction\PYGZus{}mid}\PYG{p}{(}\PYG{n+nv}{C}\PYG{p}{)} \PYG{p}{:\PYGZhy{}} \PYG{n+nf}{intelligence}\PYG{p}{(}\PYG{n+nv}{I}\PYG{p}{),}\PYG{n+nf}{ininterval}\PYG{p}{(}\PYG{n+nv}{I}\PYG{p}{,}\PYG{l+m+mi}{50}\PYG{p}{,}\PYG{l+m+mi}{60}\PYG{p}{),}
\PYG{l+s+sAtom}{\PYGZbs{}+}\PYG{n+nf}{difficulty\PYGZus{}hard}\PYG{p}{(}\PYG{n+nv}{C}\PYG{p}{).}
\PYG{n+nf}{grade\PYGZus{}high}\PYG{p}{(}\PYG{n+nv}{C}\PYG{p}{)} \PYG{p}{:\PYGZhy{}} \PYG{n+nf}{difficulty\PYGZus{}easy}\PYG{p}{(}\PYG{n+nv}{C}\PYG{p}{),} \PYG{l+s+sAtom}{\PYGZbs{}+intelligence2}\PYG{p}{,}
\PYG{l+s+sAtom}{\PYGZbs{}+}\PYG{n+nf}{nrhours2}\PYG{p}{(}\PYG{n+nv}{C}\PYG{p}{),} \PYG{l+s+sAtom}{\PYGZbs{}+intelligence1}\PYG{p}{.}
\end{Verbatim}
}
A sample program learned for the happiness data consists of rules such as:
{\scriptsize
\begin{Verbatim}[commandchars=\\\{\}]
\PYG{n+nf}{trust4}\PYG{p}{(}\PYG{n+nv}{A}\PYG{p}{)} \PYG{p}{:\PYGZhy{}} \PYG{n+nf}{trust}\PYG{p}{(}\PYG{n+nv}{A}\PYG{p}{,}\PYG{n+nv}{I}\PYG{p}{),} \PYG{n+nf}{ininterval}\PYG{p}{(}\PYG{n+nv}{I}\PYG{p}{,}\PYG{l+m+mf}{0.07857}\PYG{p}{,}  \PYG{l+m+mf}{0.1044}\PYG{p}{).}

\PYG{n+nf}{happiness1}\PYG{p}{(}\PYG{n+nv}{A}\PYG{p}{)} \PYG{p}{:\PYGZhy{}} \PYG{n+nf}{economy1}\PYG{p}{(}\PYG{n+nv}{A}\PYG{p}{),} \PYG{n+nf}{trust4}\PYG{p}{(}\PYG{n+nv}{A}\PYG{p}{).}
\PYG{n+nf}{happiness1}\PYG{p}{(}\PYG{n+nv}{A}\PYG{p}{)} \PYG{p}{:\PYGZhy{}} \PYG{n+nf}{freedom6}\PYG{p}{(}\PYG{n+nv}{A}\PYG{p}{),} \PYG{n+nf}{economy1}\PYG{p}{(}\PYG{n+nv}{A}\PYG{p}{).}
\PYG{n+nf}{happiness6}\PYG{p}{(}\PYG{n+nv}{A}\PYG{p}{)} \PYG{p}{:\PYGZhy{}} \PYG{n+nf}{health4}\PYG{p}{(}\PYG{n+nv}{A}\PYG{p}{),} \PYG{n+nf}{family2}\PYG{p}{(}\PYG{n+nv}{A}\PYG{p}{).}
\PYG{n+nf}{happiness6}\PYG{p}{(}\PYG{n+nv}{A}\PYG{p}{)} \PYG{p}{:\PYGZhy{}} \PYG{n+nf}{inregion\PYGZus{}central\PYGZus{}and\PYGZus{}eastern\PYGZus{}europe}\PYG{p}{(}\PYG{n+nv}{A}\PYG{p}{),}
\PYG{n+nf}{trust4}\PYG{p}{(}\PYG{n+nv}{A}\PYG{p}{),} \PYG{n+nf}{health3}\PYG{p}{(}\PYG{n+nv}{A}\PYG{p}{).}
\end{Verbatim}
}

\subsection{Querying}

Querying over such hybrid programs follows the syntax of the original ProbLog language. Given some evidence and a query, the success probability is calculated. Imagine, for example, that we want to determine the probability that the happiness of Slovakians is in happiness6. As Slovakia is an Eastern European country we add the evidence:
{\scriptsize 
\begin{Verbatim}[commandchars=\\\{\}]
\PYG{n+nf}{evidence}\PYG{p}{(}\PYG{n+nf}{inregion\PYGZus{}central\PYGZus{}and\PYGZus{}eastern\PYGZus{}europe}\PYG{p}{(}\PYG{l+s+sAtom}{slovakia}\PYG{p}{)).}
\end{Verbatim}
}
To formulate the query, we ask:
{\scriptsize 
\begin{Verbatim}[commandchars=\\\{\}]
\PYG{n+nf}{query}\PYG{p}{(}\PYG{n+nf}{happiness6}\PYG{p}{(}\PYG{l+s+sAtom}{slovakia}\PYG{p}{)).}
\end{Verbatim}
}
ProbLog then evaluates the query by combining the probability of happiness6 with the  corresponding rules. The {\tt ininterval} relation for each continuous relation ({\tt health4}, {\tt family2}, {\tt trust4}, {\tt health3} and {\tt happiness6}) calls an external function that calculates the integrated probability. Those probabilities are then returned to ProbLog where they are combined with the evidence and the rest of the program to generate the final probability, in this case, $0.143$.

As an illustration of multivariate continuous relations, we define a new predicate and add it to the program:
{\scriptsize 
\begin{Verbatim}[commandchars=\\\{\}]
\PYG{p}{(}\PYG{l+m+mf}{4.44} \PYG{o}{\PYGZhy{}}\PYG{l+m+mf}{17.42}\PYG{o}{*}\PYG{n+nv}{X} \PYG{o}{+} \PYG{l+m+mf}{19.66}\PYG{o}{*}\PYG{n+nv}{X}\PYG{l+s+sAtom}{\PYGZca{}}\PYG{l+m+mi}{2}\PYG{p}{)} \PYG{o}{*} \PYG{p}{(}\PYG{o}{\PYGZhy{}}\PYG{l+m+mf}{0.12}\PYG{o}{+}\PYG{l+m+mf}{0.58}\PYG{o}{*}\PYG{n+nv}{Y} \PYG{o}{+}\PYG{l+m+mf}{0.52}\PYG{o}{*}\PYG{n+nv}{Y}\PYG{l+s+sAtom}{\PYGZca{}}\PYG{l+m+mi}{2}\PYG{p}{)}\PYG{l+s+sAtom}{::}
\PYG{n+nf}{social}\PYG{p}{(}\PYG{n+nv}{X}\PYG{p}{,}\PYG{n+nv}{Y}\PYG{p}{)}
\PYG{n+nf}{social1}\PYG{o}{:\PYGZhy{}}\PYG{n+nf}{social}\PYG{p}{(}\PYG{n+nv}{X}\PYG{p}{,}\PYG{n+nv}{Y}\PYG{p}{),}\PYG{n+nf}{ininterval}\PYG{p}{(}\PYG{n+nv}{X}\PYG{p}{,}\PYG{l+m+mf}{0.4}\PYG{p}{,}\PYG{l+m+mf}{0.5}\PYG{p}{),}\PYG{n+nf}{ininterval}\PYG{p}{(}\PYG{n+nv}{Y}\PYG{p}{,}\PYG{l+m+mf}{0.42}\PYG{p}{,}\PYG{l+m+mf}{0.7}\PYG{p}{).}
\end{Verbatim}
}

We, furthermore add the new relation to the rules so that it is taken into account for our example query:
{\scriptsize 
\begin{Verbatim}[commandchars=\\\{\}]
\PYG{n+nf}{happiness6} \PYG{o}{:\PYGZhy{}} \PYG{l+s+sAtom}{health4}\PYG{p}{,} \PYG{l+s+sAtom}{family2}\PYG{p}{,} \PYG{l+s+sAtom}{social1}\PYG{p}{.}
\end{Verbatim}
}

The new program is  evaluated and returns the revised  probability 0.135.

\section{Conclusions}

To the best of our knowledge, this is the first attempt to articulate a compositional PLP framework for arbitrarily complex distributions from continuous data. It contributes an algorithmic framework that learns piecewise polynomial representations which are then related to obtain probabilistic logic programs, along with  effective symbolic inference. In our view, it takes a step towards a difficult challenge, and 
the declarative/interpretability aspect of the paradigm will be attractive for reasoning and learning over continuous data. 

\section{Acknowledgements}

This work is partly supported by the EPSRC grant `Towards Explainable and Robust Statistical AI: A Symbolic Approach'.

\bibliographystyle{aaai}
\bibliography{ijcai18}

\begin{thebibliography}{}

\bibitem[\protect\citeauthoryear{Albarghouthi \bgroup et al\mbox.\egroup
  }{2017}]{albarghouthi2017quantifying}
Albarghouthi, A.; D'Antoni, L.; Drews, S.; and Nori, A.
\newblock 2017.
\newblock Quantifying program bias.
\newblock {\em arXiv preprint arXiv:1702.05437}.

\bibitem[\protect\citeauthoryear{Baldoni \bgroup et al\mbox.\egroup
  }{2011}]{baldoni2011integrate}
Baldoni, V.; Berline, N.; De~Loera, J.; K{\"o}ppe, M.; and Vergne, M.
\newblock 2011.
\newblock How to integrate a polynomial over a simplex.
\newblock {\em Mathematics of Computation} 80(273):297--325.

\bibitem[\protect\citeauthoryear{Baldoni \bgroup et al\mbox.\egroup
  }{2014}]{baldoni2014user}
Baldoni, V.; Berline, N.; De~Loera, J.~A.; Dutra, B.; K{\"o}ppe, M.; Moreinis,
  S.; Pinto, G.; Vergne, M.; and Wu, J.
\newblock 2014.
\newblock A user's guide for latte integrale v1. 7.1.
\newblock {\em Optimization} 22:2.

\bibitem[\protect\citeauthoryear{Baral, Gelfond, and
  Rushton}{2009}]{DBLP:journals/tplp/BaralGR09}
Baral, C.; Gelfond, M.; and Rushton, J.~N.
\newblock 2009.
\newblock Probabilistic reasoning with answer sets.
\newblock {\em TPLP} 9(1):57--144.

\bibitem[\protect\citeauthoryear{Belle, Passerini, and Van~den
  Broeck}{2015}]{belle2015probabilistic}
Belle, V.; Passerini, A.; and Van~den Broeck, G.
\newblock 2015.
\newblock Probabilistic inference in hybrid domains by weighted model
  integration.
\newblock In {\em Proceedings of 24th International Joint Conference on
  Artificial Intelligence (IJCAI)},  2770--2776.

\bibitem[\protect\citeauthoryear{Bellodi and
  Riguzzi}{2011}]{bellodi2011learning}
Bellodi, E., and Riguzzi, F.
\newblock 2011.
\newblock Learning the structure of probabilistic logic programs.
\newblock In {\em International Conference on Inductive Logic Programming},
  61--75.
\newblock Springer.

\bibitem[\protect\citeauthoryear{Cerquides and
  De~M{\`a}ntaras}{1997}]{cerquides1997proposal}
Cerquides, J., and De~M{\`a}ntaras, R.~L.
\newblock 1997.
\newblock Proposal and empirical comparison of a parallelizable distance-based
  discretization method.
\newblock In {\em KDD},  139--142.

\bibitem[\protect\citeauthoryear{Chavira and
  Darwiche}{2008}]{chavira2008probabilistic}
Chavira, M., and Darwiche, A.
\newblock 2008.
\newblock On probabilistic inference by weighted model counting.
\newblock {\em Artificial Intelligence} 172(6-7):772--799.

\bibitem[\protect\citeauthoryear{Chen, Muggleton, and
  Santos}{2008}]{chen2008learning}
Chen, J.; Muggleton, S.; and Santos, J.
\newblock 2008.
\newblock Learning probabilistic logic models from probabilistic examples.
\newblock {\em Machine learning} 73(1):55--85.

\bibitem[\protect\citeauthoryear{Chistikov, Dimitrova, and
  Majumdar}{2015}]{RupakSMT}
Chistikov, D.; Dimitrova, R.; and Majumdar, R.
\newblock 2015.
\newblock Approximate counting in smt and value estimation for probabilistic
  programs.
\newblock In {\em TACAS}, volume 9035. ACTA INFORMATICA.
\newblock  320--334.

\bibitem[\protect\citeauthoryear{De~Raedt and
  Kimmig}{2015}]{de2015probabilistic}
De~Raedt, L., and Kimmig, A.
\newblock 2015.
\newblock Probabilistic (logic) programming concepts.
\newblock {\em Machine Learning} 100(1):5--47.

\bibitem[\protect\citeauthoryear{De~Raedt \bgroup et al\mbox.\egroup
  }{2015}]{de2015inducing}
De~Raedt, L.; Dries, A.; Thon, I.; Van~den Broeck, G.; and Verbeke, M.
\newblock 2015.
\newblock Inducing probabilistic relational rules from probabilistic examples.
\newblock In {\em Proceedings of 24th international joint conference on
  artificial intelligence (IJCAI)},  1835--1842.

\bibitem[\protect\citeauthoryear{De~Raedt, Kimmig, and
  Toivonen}{2007}]{de2007problog}
De~Raedt, L.; Kimmig, A.; and Toivonen, H.
\newblock 2007.
\newblock Problog: A probabilistic prolog and its application in link
  discovery.
\newblock In {\em Proceedings of the 20th international joint conference on
  Artifical intelligence},  2468--2473.

\bibitem[\protect\citeauthoryear{Deville and Lau}{1994}]{deville1994logic}
Deville, Y., and Lau, K.-K.
\newblock 1994.
\newblock Logic program synthesis.
\newblock {\em The Journal of Logic Programming} 19:321--350.

\bibitem[\protect\citeauthoryear{Dheeru and Karra~Taniskidou}{2017}]{Dua:2017}
Dheeru, D., and Karra~Taniskidou, E.
\newblock 2017.
\newblock {UCI} machine learning repository.

\bibitem[\protect\citeauthoryear{Dougherty \bgroup et al\mbox.\egroup
  }{1995}]{dougherty1995supervised}
Dougherty, J.; Kohavi, R.; Sahami, M.; et~al.
\newblock 1995.
\newblock Supervised and unsupervised discretization of continuous features.
\newblock In {\em Machine learning: proceedings of the twelfth international
  conference}, volume~12,  194--202.

\bibitem[\protect\citeauthoryear{D{\v{z}}eroski, Cestnik, and
  Petrovski}{1993}]{dvzeroski1993using}
D{\v{z}}eroski, S.; Cestnik, B.; and Petrovski, I.
\newblock 1993.
\newblock Using the m-estimate in rule induction.
\newblock {\em Journal of computing and information technology} 1(1):37--46.

\bibitem[\protect\citeauthoryear{Fierens \bgroup et al\mbox.\egroup
  }{2011}]{FierensUAI11}
Fierens, D.; {Van den Broeck}, G.; Thon, I.; Gutmann, B.; and De~Raedt, L.
\newblock 2011.
\newblock Inference in probabilistic logic programs using weighted {CNF}'s.
\newblock In {\em Proceedings of UAI},  211--220.

\bibitem[\protect\citeauthoryear{Garcia \bgroup et al\mbox.\egroup
  }{2013}]{garcia2013survey}
Garcia, S.; Luengo, J.; S{\'a}ez, J.~A.; Lopez, V.; and Herrera, F.
\newblock 2013.
\newblock A survey of discretization techniques: Taxonomy and empirical
  analysis in supervised learning.
\newblock {\em IEEE Transactions on Knowledge and Data Engineering}
  25(4):734--750.

\bibitem[\protect\citeauthoryear{Getoor \bgroup et al\mbox.\egroup
  }{2001}]{getoor2001learning}
Getoor, L.; Friedman, N.; Koller, D.; and Pfeffer, A.
\newblock 2001.
\newblock Learning probabilistic relational models.
\newblock In {\em Relational data mining}. Springer.
\newblock  307--335.

\bibitem[\protect\citeauthoryear{Gulwani}{2010}]{gulwani2010dimensions}
Gulwani, S.
\newblock 2010.
\newblock Dimensions in program synthesis.
\newblock In {\em Proceedings of the 12th international ACM SIGPLAN symposium
  on Principles and practice of declarative programming},  13--24.
\newblock ACM.

\bibitem[\protect\citeauthoryear{Gutmann \bgroup et al\mbox.\egroup
  }{2011}]{307537}
Gutmann, B.; Thon, I.; Kimmig, A.; Bruynooghe, M.; and De~Raedt, L.
\newblock 2011.
\newblock The magic of logical inference in probabilistic programming.
\newblock {\em TPLP} 11:663--680.

\bibitem[\protect\citeauthoryear{Gutmann, Jaeger, and
  De~Raedt}{2010}]{gutmann2010extending}
Gutmann, B.; Jaeger, M.; and De~Raedt, L.
\newblock 2010.
\newblock Extending problog with continuous distributions.
\newblock In {\em ILP}, volume 6489,  76--91.
\newblock Springer.

\bibitem[\protect\citeauthoryear{Gutmann, Thon, and
  De~Raedt}{2011}]{GutmannECML11}
Gutmann, B.; Thon, I.; and De~Raedt, L.
\newblock 2011.
\newblock Learning the parameters of probabilistic logic programs from
  interpretations.
\newblock In {\em ECML/PKDD (1)},  581--596.

\bibitem[\protect\citeauthoryear{Heckerman, Geiger, and
  Chickering}{1995}]{HGC95}
Heckerman, D.; Geiger, D.; and Chickering, D.
\newblock 1995.
\newblock Learning {Bayesian} networks: The combination of knowledge and
  statistical data.
\newblock {\em Machine Learning} 20:197--243.

\bibitem[\protect\citeauthoryear{Kass and Wasserman}{1995}]{kass1995reference}
Kass, R.~E., and Wasserman, L.
\newblock 1995.
\newblock A reference bayesian test for nested hypotheses and its relationship
  to the schwarz criterion.
\newblock {\em Journal of the american statistical association}
  90(431):928--934.

\bibitem[\protect\citeauthoryear{Landwehr, Kersting, and
  De~Raedt}{2005}]{landwehr2005nfoil}
Landwehr, N.; Kersting, K.; and De~Raedt, L.
\newblock 2005.
\newblock nfoil: integrating na{\"\i}ve bayes and foil.
\newblock In {\em Proceedings of the twentieth national conference on
  artificial intelligence (AAAI-05)},  795--800.

\bibitem[\protect\citeauthoryear{Lauritzen and
  Jensen}{2001}]{lauritzen2001stable}
Lauritzen, S.~L., and Jensen, F.
\newblock 2001.
\newblock Stable local computation with conditional gaussian distributions.
\newblock {\em Statistics and Computing} 11(2):191--203.

\bibitem[\protect\citeauthoryear{L{\'o}pez-Cruz, Bielza, and
  Larra{\~n}aga}{2014}]{lopez2014learning}
L{\'o}pez-Cruz, P.~L.; Bielza, C.; and Larra{\~n}aga, P.
\newblock 2014.
\newblock Learning mixtures of polynomials of multidimensional probability
  densities from data using b-spline interpolation.
\newblock {\em International Journal of Approximate Reasoning} 55(4):989--1010.

\bibitem[\protect\citeauthoryear{Milch \bgroup et al\mbox.\egroup
  }{2005}]{DBLP:conf/ijcai/MilchMRSOK05}
Milch, B.; Marthi, B.; Russell, S.~J.; Sontag, D.; Ong, D.~L.; and Kolobov, A.
\newblock 2005.
\newblock {BLOG}: Probabilistic models with unknown objects.
\newblock In {\em Proc. IJCAI},  1352--1359.

\bibitem[\protect\citeauthoryear{Muggleton}{1995}]{muggleton1995inverse}
Muggleton, S.
\newblock 1995.
\newblock Inverse entailment and progol.
\newblock {\em New generation computing} 13(3):245--286.

\bibitem[\protect\citeauthoryear{Murphy}{1999}]{murphy1999variational}
Murphy, K.~P.
\newblock 1999.
\newblock A variational approximation for bayesian networks with discrete and
  continuous latent variables.
\newblock In {\em UAI},  457--466.

\bibitem[\protect\citeauthoryear{Murphy}{2012}]{murphy2012machine}
Murphy, K.
\newblock 2012.
\newblock {\em Machine learning: a probabilistic perspective}.
\newblock The MIT Press.

\bibitem[\protect\citeauthoryear{Nitti \bgroup et al\mbox.\egroup
  }{2016}]{nitti2016learning}
Nitti, D.; Ravkic, I.; Davis, J.; and De~Raedt, L.
\newblock 2016.
\newblock Learning the structure of dynamic hybrid relational models.
\newblock In {\em 22nd European Conference on Artificial Intelligence (ECAI)
  2016}, volume 285,  1283--1290.

\bibitem[\protect\citeauthoryear{Quinlan}{1990}]{quinlan1990learning}
Quinlan, J.~R.
\newblock 1990.
\newblock Learning logical definitions from relations.
\newblock {\em Machine learning} 5(3):239--266.

\bibitem[\protect\citeauthoryear{Raedt \bgroup et al\mbox.\egroup
  }{2016}]{raedt2016statistical}
Raedt, L.~D.; Kersting, K.; Natarajan, S.; and Poole, D.
\newblock 2016.
\newblock Statistical relational artificial intelligence: Logic, probability,
  and computation.
\newblock {\em Synthesis Lectures on Artificial Intelligence and Machine
  Learning} 10(2):1--189.

\bibitem[\protect\citeauthoryear{Raghavan, Mooney, and
  Ku}{2012}]{raghavan2012learning}
Raghavan, S.; Mooney, R.~J.; and Ku, H.
\newblock 2012.
\newblock Learning to read between the lines using bayesian logic programs.
\newblock In {\em Proceedings of the 50th Annual Meeting of the Association for
  Computational Linguistics: Long Papers-Volume 1},  349--358.
\newblock Association for Computational Linguistics.

\bibitem[\protect\citeauthoryear{Ravkic, Ramon, and
  Davis}{2015}]{ravkic2015learning}
Ravkic, I.; Ramon, J.; and Davis, J.
\newblock 2015.
\newblock Learning relational dependency networks in hybrid domains.
\newblock {\em Machine Learning} 100(2-3):217--254.

\bibitem[\protect\citeauthoryear{Richardson and
  Domingos}{2006}]{richardson2006markov}
Richardson, M., and Domingos, P.
\newblock 2006.
\newblock Markov logic networks.
\newblock {\em Machine learning} 62(1):107--136.

\bibitem[\protect\citeauthoryear{Sanner and
  Abbasnejad}{2012}]{sanner2012symbolic}
Sanner, S., and Abbasnejad, E.
\newblock 2012.
\newblock Symbolic variable elimination for discrete and continuous graphical
  models.
\newblock In {\em AAAI}.

\bibitem[\protect\citeauthoryear{Schoenmackers \bgroup et al\mbox.\egroup
  }{2010}]{schoenmackers2010learning}
Schoenmackers, S.; Etzioni, O.; Weld, D.~S.; and Davis, J.
\newblock 2010.
\newblock Learning first-order horn clauses from web text.
\newblock In {\em Proceedings of the 2010 Conference on Empirical Methods in
  Natural Language Processing},  1088--1098.
\newblock Association for Computational Linguistics.

\bibitem[\protect\citeauthoryear{Shenoy and West}{2011}]{shenoy2011inference}
Shenoy, P.~P., and West, J.~C.
\newblock 2011.
\newblock Inference in hybrid bayesian networks using mixtures of polynomials.
\newblock {\em International Journal of Approximate Reasoning} 52(5):641--657.

\bibitem[\protect\citeauthoryear{Shenoy}{2012}]{shenoy2012two}
Shenoy, P.~P.
\newblock 2012.
\newblock Two issues in using mixtures of polynomials for inference in hybrid
  bayesian networks.
\newblock {\em International Journal of Approximate Reasoning} 53(5):847--866.

\bibitem[\protect\citeauthoryear{Speichert}{2017}]{Speichert:Thesis:2017}
Speichert, S.
\newblock 2017.
\newblock {Learning Hybrid Relational Rules with Piecewise Polynomial Weight
  Functions for Probabilistic Logic Programming}.
\newblock Master's thesis, The University of Edinburgh, Scotland.

\bibitem[\protect\citeauthoryear{Zettlemoyer, Pasula, and
  Kaelbling}{2005}]{zettlemoyer2005learning}
Zettlemoyer, L.~S.; Pasula, H.; and Kaelbling, L.~P.
\newblock 2005.
\newblock Learning planning rules in noisy stochastic worlds.
\newblock In {\em AAAI},  911--918.

\end{thebibliography}

\end{document}